\documentclass[letterpaper]{article} 
\usepackage{aaai25}  
\usepackage{times}  
\usepackage{helvet}  
\usepackage{courier}  
\usepackage[hyphens]{url}  
\usepackage{graphicx} 
\usepackage{natbib}  
\usepackage{caption} 
\usepackage{algorithm}

\usepackage{amsmath}
\usepackage{amssymb}
\usepackage{amsfonts}
\usepackage[misc]{ifsym}
\usepackage{color}
\usepackage{makecell}
\usepackage{soul}

\usepackage{algpseudocode}%

\usepackage{mathrsfs}
 \usepackage{bbding}
\usepackage{pifont}
\usepackage{wasysym}
\usepackage{utfsym}
\usepackage{fontawesome}
\usepackage{enumerate}
\usepackage{float}
 \usepackage{graphicx}
 \usepackage{subfig}

\usepackage{multirow, hhline}

\usepackage{xcolor}
\usepackage{colortbl, booktabs}

\definecolor{seen_back}{RGB}{224, 241, 239}
\definecolor{unseen_back}{RGB}{243, 231, 213}

\usepackage{newfloat}
\usepackage{listings}
\DeclareCaptionStyle{ruled}{labelfont=normalfont,labelsep=colon,strut=off} 
\floatstyle{ruled}
\newfloat{listing}{tb}{lst}{}
\floatname{listing}{Listing}
\title{DASK: Distribution Rehearsing via Adaptive Style Kernel Learning for Exemplar-Free Lifelong Person Re-Identification}
\author{
Kunlun Xu\textsuperscript{\rm 1}, 
Chenghao Jiang\textsuperscript{\rm 1}, 
Peixi Xiong\textsuperscript{\rm 2}, 
Yuxin Peng\textsuperscript{\rm 1}, 
Jiahuan Zhou\textsuperscript{\rm 1}\thanks{Corresponding author}
}
\affiliations{
    \textsuperscript{\rm 1}Wangxuan Institute of Computer Technology, Peking University, Beijing, China \qquad
 \textsuperscript{\rm 2}Intel Labs\\
    
xkl@stu.pku.edu.cn, cj533@scarletmail.rutgers.edu, peixi.xiong@intel.com,  \{pengyuxin, jiahuanzhou\}@pku.edu.cn
}

\usepackage{bibentry}

\begin{document}

\maketitle

\begin{abstract}
Lifelong person re-identification (LReID) is an important but challenging task that suffers from catastrophic forgetting due to significant domain gaps between training steps. Existing LReID approaches typically rely on data replay and knowledge distillation to mitigate this issue. However, data replay methods compromise data privacy by storing historical exemplars, while knowledge distillation methods suffer from limited performance due to the cumulative forgetting of undistilled knowledge. To overcome these challenges, we propose a novel paradigm that models and rehearses the distribution of the old domains to enhance knowledge consolidation during the new data learning, possessing a strong anti-forgetting capacity without storing any exemplars.
Specifically, we introduce an exemplar-free LReID method called \textbf{D}istribution Rehearsing via \textbf{A}daptive \textbf{S}tyle \textbf{K}ernel Learning (DASK). DASK includes a Distribution Rehearser Learning (DRL) mechanism that learns to transform arbitrary distribution data into the current data style at each learning step. To enhance the style transfer capacity of DRL, an Adaptive Kernel Prediction Network (AKPNet) is explored to achieve an instance-specific distribution adjustment.
Additionally, we design a Distribution Rehearsing-driven LReID Training (DRRT) module, which rehearses old distribution based on the new data via the old AKPNet model, achieving effective new-old knowledge accumulation under a joint knowledge consolidation scheme. Experimental results show our DASK outperforms the existing methods by 3.6\%-6.8\% and 4.5\%-6.5\% on anti-forgetting and generalization capacity, respectively.
Our code is available at https://github.com/zhoujiahuan1991/AAAI2025-LReID-DASK
\end{abstract}

\section{Introduction}
Person Re-identification (ReID) aims to identify a person of interest across different camera viewpoints~\cite{shi2023dual}. Extensive research has shown that ReID models trained on a specific dataset often perform poorly when applied to new datasets due to significant domain gaps~\cite{zhao2021learning, zhou2022discriminative}. This limitation has spurred a growing interest in the Lifelong Person Re-identification (LReID) task, which focuses on training models with non-stationary datasets step by step, to enhance their adaptability across various domains~\cite{pu2021lifelong}. A primary challenge in LReID is catastrophic forgetting, where the model's performance on previously learned datasets deteriorates drastically after being trained on new data~\cite{li2024exemplar}.

\begin{figure}[t]
    \begin{center}
	\includegraphics[width=1.0\linewidth]{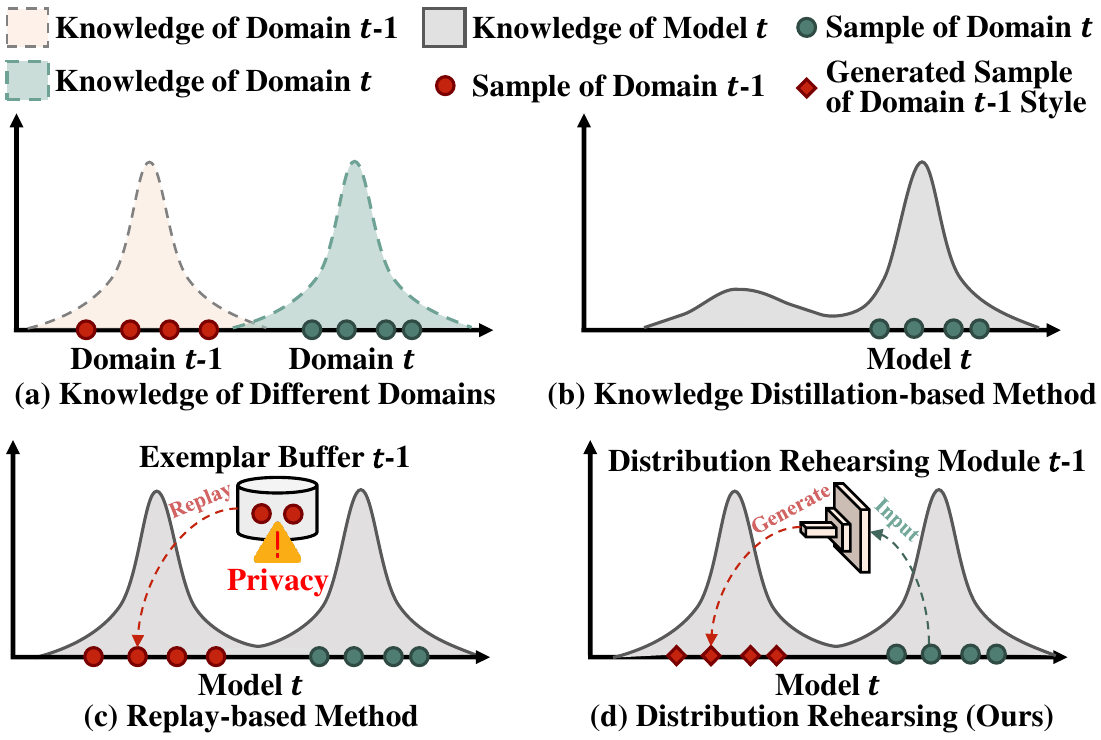}
	\caption{(a) Cross-domain distribution gap entails knowledge difference and catastrophic forgetting. (b) Knowledge distillation suffers inadequate anti-forgetting capacity.
 (c) Replay-based methods require an exemplar buffer, compromising data privacy. (d) Our method rehearses old distribution via old-style data generation, possessing a strong \textbf{anti-forgetting capacity} without storing \textbf{historical exemplars}.}
	\label{fig:motivation}
    \end{center}    
\end{figure}

Recently, various LReID methods have emerged, which can be broadly categorized into two main branches: data replay-based and knowledge distillation-based. Data replay-based methods~\cite{yu2023lifelong} rely on storing historical exemplars to rehearse old knowledge while training new models (Fig.~\ref{fig:motivation} (c)). Despite their promising anti-forgetting capabilities, these methods pose significant challenges regarding data privacy and impose a substantial storage burden, limiting their applicability in real-world scenarios~\cite{pu2022meta}. On the other hand, knowledge distillation-based methods~\cite{pu2021lifelong} focus on maintaining feature consistency between old and new models to preserve historical knowledge. However, due to the pronounced domain gap between old and new data (Fig.~\ref{fig:motivation} (a)), the features learned from historical data may not be adequately represented in the new data~\cite{xu2024lstkc}. Consequently, the corresponding discriminative knowledge is often overwritten by the new model, resulting in insufficient anti-forgetting capacity (Fig.~\ref{fig:motivation} (b)). Additionally, as undistilled historical knowledge becomes inaccessible in subsequent training steps, the forgetting effect accumulates over time, leading to progressively degraded performance on older domains.

To harness the benefits of strong anti-forgetting capabilities and privacy-friendliness from the above LReID solutions, we propose a novel paradigm that regenerates historical data distributions for old knowledge rehearsal without storing any exemplars. Our core motivation, illustrated in Fig.~\ref{fig:motivation} (d), is to develop a distribution rehearsing model capable of transforming data from the new domain to the old domain. Consequently, when new domain data is acquired, old-style data can be synthesized, enabling the model to achieve performance comparable to data replay-based methods.

To achieve this, we propose a novel LReID method named Distribution Rehearse via Adaptive Style Kernel Learning (DASK), which decomposes the LReID learning process into two key components: Distribution Rehearser Learning (DRL) and Distribution Rehearsing-driven LReID Training (DRRT). 
DRL aims to learn to transform data from arbitrary distributions to the current data style in a self-supervised manner. 
Considering that different instances possess unique offsets from the target distribution, we propose predicting a specific distribution transfer kernel for each instance utilizing an adaptive kernel prediction network~(AKPNet). Then this kernel is exploited to transform the input image into a target domain style.
DRRT aims to jointly utilize the new data and the AKPNet model obtained in the previous learning step to consolidate the new and old knowledge into a unified model. Specifically, an old AKPNet model is utilized to transform real new data into old-style data. Then, a joint knowledge consolidation module leverages the new and generated old-style data simultaneously, where a joint learning loss is employed to accumulate knowledge from both kinds of data. Extensive experimental results validate the superiority of our DASK in knowledge consolidation performance, compared to state-of-the-art exemplar-free LReID methods.

To summarize, the contributions of this paper are as follows: (1) We introduce a new LReID paradigm that rehearses the distribution of historical domains to mitigate catastrophic forgetting. (2) A novel non-exemplar LReID method named DASK is proposed, where an instance-adaptive distribution adjustment mechanism is developed to achieve high-quality old-style data generation, and a joint knowledge consolidation module is designed to achieve effective knowledge accumulation. (3) Extensive experimental results demonstrate that DASK achieves state-of-the-art performance in both anti-forgetting and generalization capacity.

\section{Related Work}
\subsection{Person Re-Identification}
Previous person re-identification (ReID) works focused on a closed setting where the test domain is identical to the training domain~\cite{shi2025multi, yin2024robust}. However, such a setting is often impractical since the domains are variable due to the dynamic environment and camera viewpoints~\cite{zhou2017efficient,zou2020joint}. 
To settle this issue, some works investigated the Domain Adaptation (DA) and Domain Generalization (DG) tasks~\cite{zheng2021exploiting,shi2024learning}. However, both DA and DG only consider utilizing static training data, neglecting the training data occurring gradually as time changes. Such a practical scenario is denoted Lifelong Person Re-Identification (LReID).  

\subsection{Lifelong Person Re-Identification}
The catastrophic forgetting problem~\cite{liu2024compositional} is the main challenge of LReID. To settle this, 
existing LReID works~\cite{ge2022lifelong,pu2023memorizing} focused on data replay and knowledge distillation strategies. 

The data reply strategy focuses on storing and replaying exemplars from historical domains~\cite{wu2021generalising, ge2022lifelong, yu2023lifelong, chen2022unsupervised, huang2022lifelong}. Although promising anti-forgetting capacity has been shown, such a strategy is impractical due to data privacy~\cite{xu2024mitigate,li2024exemplar}. Therefore, in this paper, we investigate the scenario where no historical exemplar is accessible.

Knowledge distillation~(KD)~\cite{pu2021lifelong, sun2022patch,xu2024distribution, cui2024learning} aims to mitigate the semantic drift by constraining the output consistency between the new and old model~\cite{Yang2023handling, xu2024lstkc}. 
However, since the model output can only reflect the shared features between the new and old domains, and the unique features of the historical domains are inevitably forgotten as the model updates~\cite{xu2024lstkc}, the anti-forgetting capacity of KD is usually limited. 
In this paper, to address the drawbacks of KD, we propose rehearsing the historical features from the input level.
\subsection{Distribution Rehearsing }
Existing distribution rehearsing works in ReID~\cite{deng2018image,zhou2024mixstyle,nguyen2024tackling} focused on domain adaptation and multi-domain generalization where the data of multiple training domains are available at once~\cite{li2023style,tan2023style,wei2018person}. However, in LReID, the datasets of different domains are provided step by step. Therefore, the existing methods that rely on directly utilizing multi-domain inputs are infeasible in LReID. In this paper, we instead investigate transferring the data of arbitrary distributions to the known style, thereby the old distribution can be rehearsed without storing historical exemplars. 

\begin{figure*}[t]
		\begin{center}
			\includegraphics[width=0.99\linewidth]{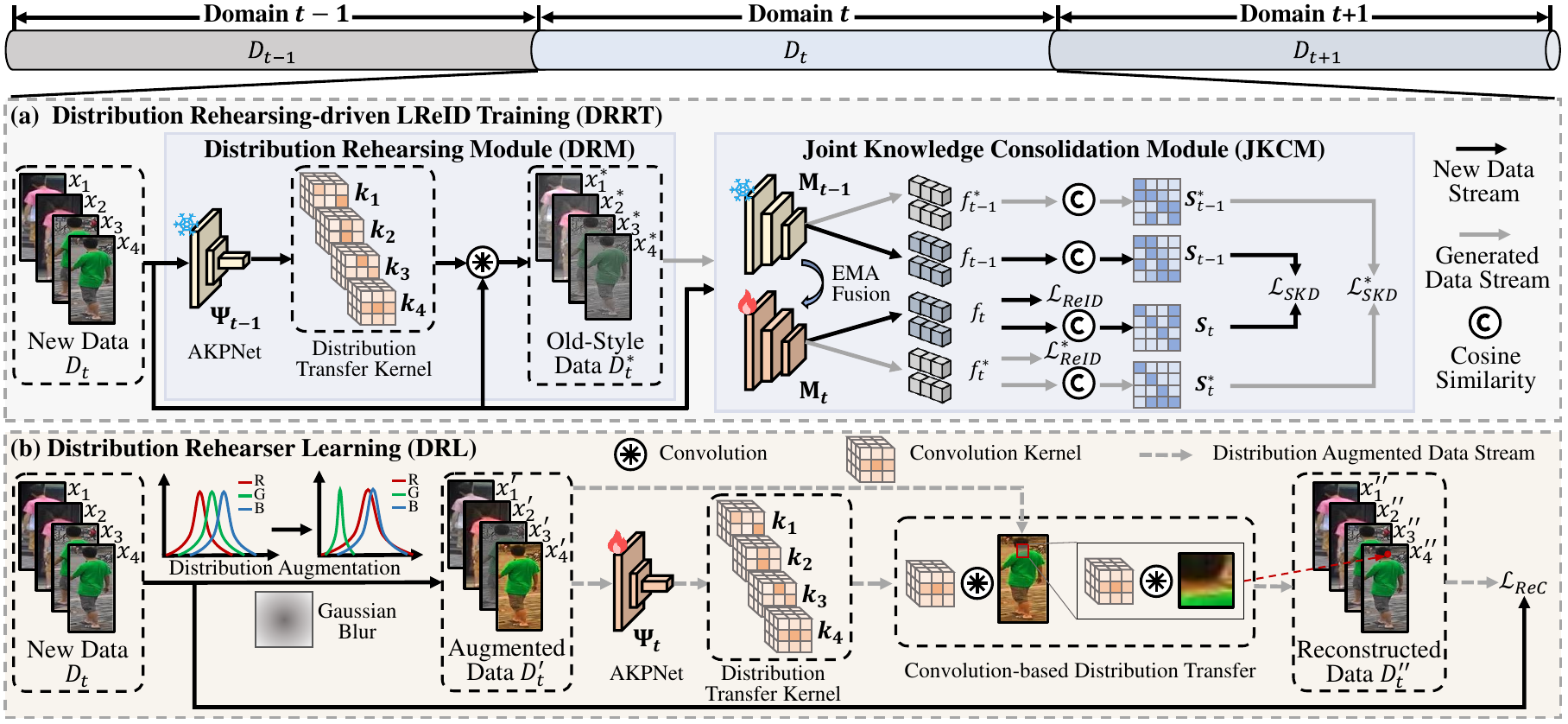}
			\caption{The overall components of our DASK method. At each training step, a new domain dataset $D_t$ is given. (a) 
   DRRT scheme generates the old-style data to enhance knowledge consolidation as the LReID model $\boldsymbol{\mathrm{M}}_t$ learns from $D_t$. 
   (b) DRL mechanism trains an Adaptive Kernel Prediction Network (AKPNet) to achieve instance-specific distribution adjustment, aiming to transform the data of arbitrary domains to domain $t$, preparing for $D_t$-style data generation in subsequent training steps.}
			\label{fig:framework}
		\end{center}    
	\end{figure*}
\section{The Proposed Method}
\subsection{Problem Definition}
In the LReID task, a stream of $T$ training datasets $\mathcal{D}=\{D_t\}_{t=1}^{T}$ is given step by step, where $D_t=\{(x_i,y_i)\}_{i=1}^{n_t}$ contains $n_t$ pairs of image $x_i$ and corresponding identity label $y_i$. When $D_t$ is given, the previous $t-1$ datasets are inaccessible~\cite{xu2024distribution}. After the $t$-th training step, the trained model is denoted as $\boldsymbol{\mathrm{M}}_t$. The final model $\boldsymbol{\mathrm{M}}_T$ is evaluated on $T$ testing datasets $\mathcal{D}^{te}=\{D_t^{te}\}_{t=1}^{T}$ which are collected form $T$ training domains. Besides, additional $U$ datasets $\mathcal{D}^{un}=\{ D^{un}_t\}_{t=1}^{U}$ collected from novel domains are used to evaluate the generalization capacity $\boldsymbol{\mathrm{M}}_T$.
\subsection{Overview}
As shown in Fig.~\ref{fig:framework}, the proposed method contains a Distribution Rehearsing-driven LReID Training (DRRT) scheme and an additional Distribution Rehearser Learning (DRL) mechanism. DRRT aims to learn a ReID model $\boldsymbol{\mathrm{M}}_t$ based on the old model $\boldsymbol{\mathrm{M}}_{t-1}$, an new AKPNet model $\boldsymbol{\mathrm{\Psi}}_{t-1}$ and the new data $D_t$. DRL aims to obtain an AKPNet model $\boldsymbol{\mathrm{\Psi}}_{t}$ of the current training step based on the new data $D_t$.

\subsection{Distribution Rehearsing-driven LReID Training}
The main function of DRRT is to generate old-style data and jointly utilize the real new data and generated data to achieve new knowledge learning and mitigate catastrophic forgetting.
As shown in Fig.~\ref{fig:framework} (a), DRRT primarily contains a Distribution Rehearsing Module (DRM) and a Joint Knowledge Consolidation Module (JKCM).

\textbf{Distribution Rehearsing Module}:
Given the new data $D_t$, DRM utilizes an AKPNet model $\boldsymbol{\mathrm{\Psi}}_{t-1}$ to generate a distribution transfer kernel $\boldsymbol{k}_i\in\mathbb{R}^{C\times C\times k \times k}$ for each image $x_i\in\mathbb{R}^{W\times H \times C}$, \textit{i.e.} $\boldsymbol{k}_i=\boldsymbol{\mathrm{\Psi}}_{t-1}(x_i)$, where $H$, $W$, and $C$ are the image height, width, and channel, respectively. $\boldsymbol{k}_i$ is actually a convolution kernel and $k$ is the kernel size. Then, $\boldsymbol{k}_i$ is utilized to process $x_i$ by:
\begin{equation}
    x_i^*=\boldsymbol{k}_i\circledast x_i
    \label{eq:synthetic_t-1},
\end{equation}
where $\circledast$ denotes the convolution process and $x_i^*\in\mathbb{R}^{W\times H \times C}$ is a generated image that contains the distribution information of the historical domain. Then, $x_i^*$ is assigned the same identity label $y_i$ with $x_i$.

\textbf{Joint Knowledge Consolidation Module}:
When the real new data $D_t$ and the generated data $D_t^*$ are obtained, a frozen old ReID model $\boldsymbol{\mathrm{M}}_{t-1}$ and a learnable new ReID model $\boldsymbol{\mathrm{M}}_{t}$ are utilized to process $D_t$ and $D_t^*$, where the parameters of $\boldsymbol{\mathrm{M}}_{t}$ are initialized with $\boldsymbol{\mathrm{M}}_{t-1}$. 

For the new data $D_t$, given a batch of $B$ input images $\boldsymbol{x}=\{x_i\}_{i=1}^{B}$, the ReID models transform each image into a $d$-dimensional feature.
Specifically, the features extracted by $\boldsymbol{\mathrm{M}}_{t-1}$ and $\boldsymbol{\mathrm{M}}_{t}$ are denoted as  $\boldsymbol{f}_{t-1}=\{f_{t-1}^{(i)}\in\mathbb{R}^d\}_{i=1}^B$ and $\boldsymbol{f}_{t}=\{f_{t}^{(i)}\in\mathbb{R}^d\}_{i=1}^B$, separately. 
Then, the cross-instance similarity matrices $\boldsymbol{S}_{t-1}\in\mathbb{R}^{B\times B}$ and $\boldsymbol{S}_{t}\in\mathbb{R}^{B\times B}$ are calculated by :
\begin{equation}     
    \left\{
		\begin{aligned}
			  \boldsymbol{S}_{t-1}&=\frac{F_{t-1}}{||F_{t-1}||} \cdot\frac{F_{t-1}^\top}{||F_{t-1}^\top||}\\		 \boldsymbol{S}_{t}&=\frac{F_{t}}{||F_{t}||} \cdot\frac{F_{t}^\top}{||F_{t}^\top||}
		\end{aligned}
		\right.,
    \label{eq:cosine_similarity}
\end{equation}
where $F_{t-1}\in\mathbb{R}^{B\times d}$ and $F_t\in\mathbb{R}^{B\times d}$ are the matrix forms of $\boldsymbol{f}_{t-1}$ and $\boldsymbol{f}_{t}$, respectively. $\frac{F_{t-1}}{||F_{t-1}||}$ is a row-wise L2 normalization process to facilitate model convergence \cite{xu2024distribution}. Following \cite{xu2024lstkc}, the cross-instance similarity knowledge distillation loss is adopted:
\begin{equation}
   \mathcal{L}_{SKD}=\frac{1}{B}\sum_{i=1}^{B} KL((\boldsymbol{S}_{t-1})_i \big\vert\big\vert (\boldsymbol{S}_t)_i )
    \label{eq:KL_loss},
\end{equation}
where $(\boldsymbol{S}_{t})_i$ denotes the $i$-th row of matrix $\boldsymbol{S}_t$. $\mathcal{L}_{SKD}$ is an anti-forgetting loss that aims to maintain the predicted cross-instance similarity between the old and new models.

To ensure the new knowledge learning, the classical ReID loss $\mathcal{L}_{ReID}$ is adopted which consists of a Triplet loss $\mathcal{L}_{Tri}$~\cite{he2021transreid} and a Cross-Entropy loss $\mathcal{L}_{CE}$. Following the previous LReID works \cite{xu2024distribution}, $\mathcal{L}_{ReID}$ is calculated by:
\begin{equation}
   \mathcal{L}_{ReID}=\mathcal{L}_{Tri}+\mathcal{L}_{CE}
    \label{eq:reid_loss}.
\end{equation}

 Besides, for the generated data $D_t^*$,  $\boldsymbol{\mathrm{M}}_{t-1}$ and $\boldsymbol{\mathrm{M}}_{t}$ are adopted to extract the features $\boldsymbol{f}_{t-1}^*=\{(f_{t-1}^{(i)})^*\in\mathbb{R}^d\}_{i=1}^B$ and $\boldsymbol{f}_{t}^*=\{(f_{t}^{(i)})^*\in\mathbb{R}^d\}_{i=1}^B$ from the input batch $\boldsymbol{x}^*=\{x_i^*\}_{i=1}^{B}$. Then, Eq.~\ref{eq:cosine_similarity} and Eq. \ref{eq:KL_loss} are adopted to obtain rehearsed SKD loss $\mathcal{L}_{SKD}^*$ based on $\boldsymbol{f}_t^*$ and $\boldsymbol{f}_{t-1}^*$. 
 Then, the rehearsed ReID loss 
  $\mathcal{L}_{ReID}^*$ is calculated following Eq.~\ref{eq:reid_loss}.

Note that $\mathcal{L}_{ReID}^*$ ensures the new model $\boldsymbol{\mathrm{M}}_{t}$ learns with old-style data under accurate new identity labels, which not only enhance historical discriminative knowledge consolidation but also guide the model to mine the domain-irrelevant knowledge, thus improving the model's generalization capability. Besides, $\mathcal{L}_{SKD}^*$ aims to maintain the cross-instance similarity structure under the historical data style, improving the structural knowledge transfer from $\boldsymbol{\mathrm{M}}_{t-1}$ to $\boldsymbol{\mathrm{M}}_{t}$, which is complementary to 
 the function of $\mathcal{L}_{ReID}^*$.

\textbf{ReID Model Training and Inference}: When jointly utilizing $D_t^*$ and $D_t$ to training the new model $\boldsymbol{\mathrm{M}}_{t}$, the overall loss is calculated by:
 \begin{equation}
   \mathcal{L}_{Total}=\mathcal{L}_{ReID} +\alpha \mathcal{L}_{SKD}+\beta(\mathcal{L}_{ReID}^* +\alpha \mathcal{L}_{SKD}^*)
    \label{eq:overall_loss},
\end{equation}
where $\alpha$ and $\beta$ are hyperparameters to balance the new knowledge learning and historical knowledge forgetting. Besides, following previous works \cite{li2024exemplar, xu2024distribution}, the Exponential Moving Average (EMA) strategy is adopted to fuse the new and old model:
 \begin{equation}
   \boldsymbol{\mathrm{M}}_{t}\leftarrow \lambda\boldsymbol{\mathrm{M}}_{t-1} +(1-\lambda)\boldsymbol{\mathrm{M}}_{t}
    \label{eq:ema},
\end{equation}
where $\lambda$ is a weight to balance the new and old parameters.

During inference, only the final model $\boldsymbol{\mathrm{M}}_{T}$ is utilized to extract image features for person matching. Thus our method does not introduce additional computational overhead beyond a feature extractor backbone at the test stage.

\subsection{Distribution Rehearser Learning }
Due to data privacy, the data of different domains can not be obtained simultaneously, thus the existing style transfer methods that rely on the data of multiple accessible domains are infeasible~\cite{deng2018image,zhou2024mixstyle,nguyen2024tackling}. To settle this, in our DRL mechanism, we first introduce a \textbf{Distribution Augmentation} strategy to generate random domain data for cross-domain distribution transfer learning. Then, a \textbf{Distribution Reconstruction} design is adopted to guide the model to learn to transfer the distribution augmented data into the current domain.

\textbf{Distribution Augmentation}:
Firstly, given the new domain data $D_t$, we obtain the mean $\boldsymbol{\mu}_i=\{\mu_{i}^r,\mu_{i}^g,\mu_{i}^b\}$ and standard deviation $\boldsymbol{\sigma}_i=\{\sigma_{i}^r,\sigma_{i}^g,\sigma_{i}^b\}$ of the $R, G, B$ channels in each image $x_i$. 
Then, the standard deviations of $\boldsymbol{\mu}_i$ and $\boldsymbol{\sigma}_i$ across the images in $D_t$ are also obtained, which are denoted as $\overline{\boldsymbol{\sigma}}_\mu=\{\overline{\sigma}_{\mu}^r,\overline{\sigma}_{\mu}^g,\overline{\sigma}_{\mu}^b\}$ and $\overline{\boldsymbol{\sigma}}_\sigma=\{\overline{\sigma}_{\sigma}^r,\overline{\sigma}_{\sigma}^g,\overline{\sigma}_{\sigma}^b\}$, respectively. 
Note that $\boldsymbol{\mu}_i$ and $\boldsymbol{\sigma}_i$ reflect the instance-specific color distribution, while $\overline{\boldsymbol{\sigma}}_\mu$ and $\overline{\boldsymbol{\sigma}}_\sigma$ reflect the overall color distribution within domain $t$.

Then, for an input image $x_i$, its distribution is augmented by Gaussian Sampling. Specifically, the augmented mean ${\mu_i^r}'$ and standard deviation ${\sigma_i^r}'$ of the $R$ channel are sampled according to 
${\mu_i^r}'\sim \mathcal{N}(\mu_i^r, {\overline{\sigma}_\mu^r}^2)$ and ${\sigma_i^r}'\sim \mathcal{N}(\sigma_i, {\overline{\sigma}_\sigma^r}^2)$, respectively. Furthermore, the augmented $R$ channel data ${x_i^r}'\in\mathbb{R}^{W\times H}$ of the image $x_i$ is obtained by:
 \begin{equation}
   ({x_i^r}')_{m,n}=\frac{(x_i^r)_{m,n}-\mu_i^r+{\mu_i^r}'}{\sigma_i^r}\cdot{\sigma_i^r}'
    \label{eq:channel-aug},
\end{equation}
where $x_i^r$ is the original $R$ channel data of $x_i$, $(m,n)$ is the coordinate of a pixel. Similarly, the augmented $G$ channel ${x_i^g}'$ and $B$ channel ${x_i^b}'$ are generated.
Once the color distribution augmentation is applied, the random Gaussian blur is also employed to further improve the texture diversity of the augmented image ${x_i}'$. 

\textbf{Distribution Reconstruction}:
To address the factor that different instances possess unique distribution drifts, we introduce the adaptive kernel prediction network (AKPNet) which consists of a lightweight CNN backbone and a linear layer. Specifically, AKPNet generates a Distribution Transfer Kernel $\boldsymbol{k}_i$ for the input $x_i'$, as shown in Fig.~\ref{fig:framework} (b). Then, $\boldsymbol{k}_i$ serves as a convolution kernel to process $x_i'$ and the obtained reconstructed image is denoted as $x_i''$. A self-supervised reconstruction loss $\mathcal{L}_{ReC}$ is applied as follows:
 \begin{equation}
   \mathcal{L}_{ReC}=||x_i-x_i''||
    \label{eq:loss-reconstruct},
\end{equation}
where $x_i''=\boldsymbol{\mathrm{\Psi}}_t(x_i')\circledast x_i'$. Therefore, $\boldsymbol{\mathrm{\Psi}}_t$ is forced to recover the original distribution of $x_i$ from the augmented data $x_i'$. Thus when the new data $D_{t+1}$ from an arbitrary domain is given, the distribution of $D_t$ can be rehearsed.

\begin{table*}[htbp]
   \centering
  \setlength{\tabcolsep}{0.75mm}{
    \begin{tabular}{llccccccccccc>{\columncolor{seen_back}}c>{\columncolor{seen_back}}c>{\columncolor{unseen_back}}c>{\columncolor{unseen_back}}cccc}   
    \hline
   &\multirow{2}[1]{*}{Method}&\multirow{2}[1]{*}{Publication} & \multicolumn{2}{c}{Market-1501} & \multicolumn{2}{c}{CUHK-SYSU} & \multicolumn{2}{c}{DukeMTMC} & \multicolumn{2}{c}{MSMT17} & \multicolumn{2}{c}{CUHK03} & \multicolumn{2}{>{\columncolor{seen_back}}c}{\textbf{Seen-Avg}} & \multicolumn{2}{>{\columncolor{unseen_back}}c}{\textbf{UnSeen-Avg}} \\
\hhline{~~~--------------}
&&& mAP   & R@1   & mAP   & R@1   & mAP   & R@1   & mAP   & R@1   & mAP   & R@1   & mAP   & R@1   & mAP   & R@1 \\
\hline

&JointTrain   &-  &78.9  & 90.9  & 86.7  & 88.2  & 71.2  & 82.9  & 36.2  & 61.2  & 61.2  & 63.4  & 66.8  & 77.3  & 59.4  & 52.6  \\

\hline
\multirow{5}[0]{*}{\rotatebox{90}{CIL} } 
&LwF &T-PAMI 2017             & 56.3  & 77.1  & 72.9  & 75.1  & 29.6  & 46.5  & 6.0   & 16.6  & 36.1  & 37.5  & 40.2  & 50.6  & 47.2  & 42.6  \\ 
&SPD &ICCV 2019      & 35.6  & 61.2  & 61.7  & 64.0  & 27.5  & 47.1  & 5.2   & 15.5  & 42.2  &44.3  & 34.4  & 46.4  & 40.4  & 36.6  \\  
&PRD&ICML 2023   & 7.3   & 18.0  & 33.5  & 35.6  & 3.7   & 7.6   & 0.8   & 2.4   & 33.8  & 33.8  & 15.8  & 19.5  & 23.0  & 17.7  \\
&PRAKA& ICCV 2023   & 37.4  & 61.3  & 69.3  & 71.8  & 35.4  & 55.0  & 10.7  & 27.2  & \textcolor{red}{\textbf{54.0}}  & \textcolor{red}{\textbf{55.6}}  & 41.3  & 54.2  & 47.7  & 41.6  \\
&FCS &CVPR 2024            &58.3  & 78.5  & 75.1  & 76.2  & 42.6  & 59.8  & 10.2  & 24.3  & 35.3  & 34.9  & 44.3  & 54.7  & 52.1  & 44.2  \\

\hline
\multirow{9}[0]{*}{\rotatebox{90}{LReID} } 
&CRL &WACV 2021          & 58.0  & 78.2  & 72.5  & 75.1  & 28.3  & 45.2  & 6.0   & 15.8  & 37.4  & 39.8  & 40.5  & 50.8  & 47.8  & 43.5  \\
    &AKA &CVPR 2021  & 51.2  & 72.0  & 47.5  & 45.1  & 18.7  & 33.1  & 16.4  & 37.6  & 27.7  & 27.6  & 32.3  & 43.1  & 44.3  & 40.4  \\
&
PatchKD & ACM MM 2022           &\textcolor{red}{\textbf{68.5}}  & \textcolor{red}{\textbf{85.7}}  & 75.6  & 78.6  & 33.8  & 50.4  & 6.5   & 17.0  & 34.1  & 36.8  &43.7  &53.7  &49.1  &45.4  \\    
&MEGE&T-PAMI 2023  & 39.0    & 61.6  & 73.3  & 76.6  & 16.9  & 30.3  & 4.6   & 13.4  & 36.4  & 37.1  & 34.0    & 43.8  & 47.7  & 44.0 \\
&ConRFL &PR 2023      & 59.2  & 78.3  &\textcolor{blue}{\textbf{82.1}}  &\textcolor{blue}{\textbf{84.3}}  & 45.6  & 61.8  & 12.6  & 30.4  &\textcolor{blue}{\textbf{51.7}}  &\textcolor{blue}{\textbf{53.8}} & 50.2  & 61.7  & 57.4  &\textcolor{blue}{\textbf{52.3}} \\
&CKP& ACM MM 2024&53.8  & 76.0  & 81.2  & 83.0  &49.7  &67.0  & 18.4  & 40.8  & 44.1  & 45.8  & 49.4  & 62.5  &58.0  & 51.0\\
&LSTKC &AAAI 2024&54.7  & 76.0  &81.1  & 83.4  & 49.4  &66.2  & \textcolor{blue}{\textbf{20.0}}  & \textcolor{blue}{\textbf{43.2}}  &44.7  &46.5  &50.0  &63.1  &57.0  &49.9  \\
&DKP &CVPR 2024         &60.3	&80.6 & \textcolor{red}{\textbf{83.6}} & \textcolor{red}{\textbf{85.4}} & \textcolor{blue}{\textbf{51.6}} & \textcolor{blue}{\textbf{68.4}} &19.7 &41.8&43.6&44.2 & \textcolor{blue}{\textbf{51.8}} & \textcolor{blue}{\textbf{64.1}} & \textcolor{blue}{\textbf{59.2}} & 51.6\\	
\hhline{~----------------}
&\textbf{DASK} & This Paper&\textcolor{blue}{\textbf{61.2}}  &\textcolor{blue}{\textbf{82.3}}  & 81.9  & 83.7  &\textcolor{red}{\textbf{58.5}}  &\textcolor{red}{\textbf{74.6}}  &\textcolor{red}{\textbf{29.1}} &\textcolor{red}{\textbf{ 57.6}}  & 46.2  & 48.1  &\textcolor{red}{\textbf{55.4}}  & \textcolor{red}{\textbf{69.3}}  &\textcolor{red}{\textbf{65.3}}  &\textcolor{red}{\textbf{58.4}}  \\
    \hline  
    \end{tabular}%
   }
 \raggedright
\caption{
Comparison results of Seen-domain anti-forgetting and UnSeen-domain generalization on Training Order-1.
  }
\label{tab:setting1}%
\end{table*}%

\subsection{Discussion and Analysis}
In this section, we discuss the effectiveness of DASK in learning to transform $x_i'$ into the distribution of $x_i$, in comparison with existing distribution rehearsing solutions.

\textbf{Compared to Statistics Predicting-based Method}: The existing method, CoP
 \cite{gu2023color} considers the reverse process of Eq.~\ref{eq:channel-aug}, where the intensity of the $R$ channel of a pixel $(x_i^r)_{m,n}$ can be represented as:
\begin{equation}
  {(x_i^r)_{m,n}}=\frac{\sigma_i^r}{{\sigma_i^r}'} ({x_i^r}')_{m,n}+\mu_i^r-{\mu_i^r}'
    \label{eq:reverse-channel-aug},
\end{equation}
which is a hypothetical model that only involves the linear transformation of pixels. However, directly predicting $\frac{\sigma_i^r}{{\sigma_i^r}'}$ and $\mu_i^r-{\mu_i^r}'$ lacks the content-level constraints, leading to unrealistic reconstruction as shown in Fig.~\ref{fig:feature-distribution} (b). Instead, in our DASK,
the reconstruction for each pixel $(x_i^r)_{m,n}$ can be deduced as a neighborhood-weighting process:
\begin{equation}
\small
  {(x_i^r)_{m,n}}=\sum_{j\in\{r,g,b\}}\sum_{p=-1}^1\sum_{q=-1}^1 k^j_{i,p,q}({x_i^r}')_{m+p,n+q}+b^r_i
    \label{eq:reverse-channel-aug-conv},
\end{equation}
where $k^j_{i,p,q}$ is a weight element of $\boldsymbol{k}_i$, and $b^r_i$ is the bias of channel $R$. Note that Eq.~\ref{eq:reverse-channel-aug} is a special case of Eq.~\ref{eq:reverse-channel-aug-conv} when $k^j_{i,p,q}$=0, $w.r.t.\thinspace p\neq 0, q\neq 0$. Thus, the proposed instance-adaptive transfer kernel prediction can handle the instance-specific linear hypothesis and model the adjacent relations simultaneously, obtaining better reconstructions.

\textbf{Compared to Shared Generation Network-based Methods}:
When training a shared generation network for the images, if a shallow network is adopted, the parameter $k^j_{i,p,q}$ in Eq.~\ref{eq:reverse-channel-aug-conv} is fixed across all inputs. Thus the reconstruction loss forces the model to learn an average style of the target domain (as the illustration of Shared Conv in Fig.~\ref{fig:feature-distribution} (b)). While deep networks can model more complex distributions, they demand large amounts of training data, which is impractical for lifelong learning~\cite{wang2020collaborative}. Besides, deep generation networks introduce higher computational overhead~\cite{wang2023microast} compared to our lightweight AKPNet. Therefore, our approach offers an effective and efficient means to achieve high-quality reconstruction.

\section{Experiment}
\subsection{Datasets and Evaluation Metrics}
\textbf{Datasets}: All the experiments are conducted on the LReID benchmark \cite{pu2021lifelong}, which contains five training subsets (Market-1501 \cite{zheng2015scalable}, DukeMTMC-reID \cite{ristani2016performance}, CUHK-SYSU \cite{xiao2016end}, MSMT17-V2 \cite{wei2018person}, and CUHK03~\cite{li2014deepreid}) and seven novel subsets (CUHK01~\cite{li2012human}, CUHK02~\cite{li2013locally}, VIPeR~\cite{gray2008viewpoint}, PRID~\cite{hirzer2011person}, i-LIDS~\cite{branch2006imagery}, GRID~\cite{loy2010time}, and SenseReID~\cite{zhao2017spindle}). The training and novel subsets are used for anti-forgetting and generalization evaluation, respectively. Following the previous works \cite{sun2022patch}, two training orders are adopted to imitate varying domain gaps~\footnote{(Order-1) Market-1501$\rightarrow$CUHK-SYSU$\rightarrow$DukeMTMC-reID $\rightarrow$ MSMT17$\rightarrow$CUHK03}~\footnote{(Order-2) DukeMTMC-reID$\rightarrow$MSMT17$\rightarrow$Market-1501$\rightarrow$ CUHK-SYSU $\rightarrow$CUHK03}.

\textbf{Evaluation Metrics}: Following previous LReID works \cite{pu2021lifelong, sun2022patch}, the mean Average Precision (mAP) and Rank@1 (R@1) accuracy are calculated on the subsets to show the model's adaptation to each domain. Besides, the average mAP and R@1 are also calculated to evaluate the overall anti-forgetting and generalization capacity on the training (Seen) and novel (Unseen) domains.

\subsection{Implementation Details}
To ensure a fair comparison with previous works~\cite{xu2024distribution}, we adopt ResNet-50 as the backbone of our ReID model. The backbone of the AKPNet is the Mobilenet-v3~\cite{howard2019searching}. When training the ReID model, the first dataset $D_1$ is trained for 80 epochs and the subsequent $t-1$ datasets are trained for 60 epochs. As for the distribution rehearsing learning, each dataset is trained for 50 epochs.
The input images are resized to $256 \times 128$ with random cropping, erasing, and horizontal flipping augmentation. 
The hyperparameters $\alpha$, $\beta$, and $\lambda$ are set to 1.0, 4.5, and 0.5 respectively.  The training pipeline shown in Fig. \ref{fig:framework} (a) DRRT and (b) DRL are conducted independently.
To prevent the continuous growth of storage overhead, \textbf{only one old AKPNet model}, $\boldsymbol{\mathrm{\Psi}}_{t-1}$, is retained for DRRT at the $t$-th training step. All experiments are conducted on a single NVIDIA 3090 GPU.

\begin{table*}[htbp]
  \centering
  \setlength{\tabcolsep}{0.75mm}{
    \begin{tabular}{llccccccccccc>{\columncolor{seen_back}}c>{\columncolor{seen_back}}c>{\columncolor{unseen_back}}c>{\columncolor{unseen_back}}cccc}
    \hline
    &\multirow{2}[1]{*}{Method}&\multirow{2}[1]{*}{Publication}& \multicolumn{2}{c}{DukeMTMC} & \multicolumn{2}{c}{MSMT17} & \multicolumn{2}{c}{Market-1501} & \multicolumn{2}{c}{CUHK-SYSU} & \multicolumn{2}{c}{CUHK03} & \multicolumn{2}{>{\columncolor{seen_back}}c}{\textbf{Seen-Avg}} & \multicolumn{2}{>{\columncolor{unseen_back}}c}{\textbf{UnSeen-Avg}} \\
\hhline{~~~--------------}
&& & mAP   & R@1   & mAP   & R@1   & mAP   & R@1   & mAP   & R@1   & mAP   & R@1   & mAP   & R@1   & mAP   & R@1 \\
\hline   
    &JointTrain &-& 71.2  & 82.9  & 36.2  & 61.2  & 78.9  & 90.9  & 86.7  & 88.2  & 61.2  & 63.4  & 66.8  & 77.3  & 59.4  & 52.6  \\
    \hline
\multirow{5}[0]{*}{\rotatebox{90}{CIL} }   
    &LwF &T-PAMI 2017      & 42.7  & 61.7  & 5.1   & 14.3  & 34.4  & 58.6  & 69.9  & 73.0  & 34.1  & 34.1  & 37.2  & 48.4  & 44.0  & 40.1  \\
    &SPD& ICCV 2019  & 28.5  & 48.5  & 3.7   & 11.5  & 32.3  & 57.4  & 62.1  & 65.0  &43.0  & 45.2  & 33.9  & 45.5  & 39.8  & 36.3  \\
    &PRD & ICML 2023  & 3.6   & 8.2   & 0.6   & 1.8   & 8.9   & 22.3  & 34.6  & 36.1  & 35.4  & 35.3  & 16.6  & 20.7  & 20.7  & 15.0  \\
    &PRAKA & ICCV 2023  & 31.2  & 48.7  &6.6   &19.1  &47.8  &69.8  & 70.4  & 73.0  & \textcolor{red}{\textbf{54.9}}  & \textcolor{red}{\textbf{56.6}}  & 42.2  & 53.4  & 48.4  & 41.1  \\    
    &FCS   &  CVPR 2024   &53.6  & 70.0  & 9.5   & 23.5  & 48.7  & 69.9  & 76.2  & 78.2  & 37.1  & 38.4  & 45.0  & 56.0  & 52.7  & 45.1  \\
    \hline
\multirow{9}[0]{*}{\rotatebox{90}{LReID} } 
    &CRL &  WACV 2021    & 43.5  & 63.1  & 4.8   & 13.7  & 35.0  & 59.8  & 70.0  & 72.8  & 34.5  & 36.8  & 37.6  & 49.2  & 45.3  & 41.4  \\
    &AKA & CVPR 2021  & 32.5  & 49.7  & -     & -     & -     & -     & -     & -     & -     & -     & -     & -     & 40.8  & 37.2  \\
    &PatchKD & ACM MM 2022         &\textcolor{red}{\textbf{58.3}}  &\textcolor{blue}{\textbf{74.1}}  & 6.4   & 17.4  & 43.2  & 67.4  &74.5  &76.9  & 33.7  & 34.8  &43.2  &54.1  &48.6  &44.1  \\  
    &MEGE& T-PAMI 2023  & 21.6  & 35.5  & 3.0     & 9.3   & 25.0    & 49.8  & 69.9  & 73.1  & 34.7  & 35.1  & 30.8  & 40.6  & 44.3  & 41.1 \\ 
    &ConRFL& PR 2023 &34.4  & 51.3  & 7.6   & 20.1  &\textcolor{blue}{\textbf{61.6}}  & 80.4  & 82.8  &\textcolor{blue}{\textbf{85.1}}  &\textcolor{blue}{\textbf{49.0}}    &\textcolor{blue}{\textbf{50.1}}  & 47.1  & 57.4  & 57.9  &\textcolor{blue}{\textbf{53.4}}\\       
    &CKP&ACM MM 2024&49.4  & 67.0  & 14.5  & 33.8  & 56.0  & 77.6  &\textcolor{blue}{\textbf{83.2}}  & 84.9  & 45.3  & 47.1  & 49.7  & 62.1  & 57.2  & 50.0  \\
    &LSTKC& AAAI 2024 & 49.9  & 67.6  &\textcolor{blue}{\textbf{14.6}}  &\textcolor{blue}{\textbf{34.0}}  &55.1  &76.7  &82.3  &83.8  &46.3  &48.1  &49.6  &62.1  &57.6  &49.6  \\    
    &DKP& CVPR 2024 &53.4	&70.5	&14.5	&33.3	&60.6	&\textcolor{blue}{\textbf{81.0}}	&83.0	&84.9	&45.0	&46.1	&\textcolor{blue}{\textbf{51.3}}	&\textcolor{blue}{\textbf{63.2}}    &\textcolor{blue}{\textbf{59.0}}	&51.6\\
    \hhline{~----------------}
&\textbf{DASK} & This Paper&\textcolor{blue}{\textbf{55.7}}  &\textcolor{red}{\textbf{74.4}}  & \textcolor{red}{\textbf{25.2}}  & \textcolor{red}{\textbf{51.9}}  & \textcolor{red}{\textbf{71.6}}  & \textcolor{red}{\textbf{87.7}}  & \textcolor{red}{\textbf{84.8}}  & \textcolor{red}{\textbf{86.2}}  & 48.4  & 49.8  & \textcolor{red}{\textbf{57.1}}  & \textcolor{red}{\textbf{70.0}}  & \textcolor{red}{\textbf{65.5}}  & \textcolor{red}{\textbf{57.9}}  \\
\hline
    \end{tabular}%
   }
  \caption{
Comparison results of Seen-domain anti-forgetting and UnSeen-domain generalization on Training Order-2.
  }
  \label{tab:setting2}%
\end{table*}%

\subsection{Compared Methods}
We compare the proposed method with two streams of methods: \textit{Class Incremental Learning (CIL)} methods (LwF~\cite{li2017learning}, SPD~\cite{tung2019similarity}, 
PRAKA~\cite{shi2023prototype}, and FCS~\cite{li2024fcs}) and \textit{Non-exemplar LReID} methods (CRL~\cite{zhao2021continual}, AKA~\cite{pu2021lifelong}, PatchKD~\cite{sun2022patch}, ConRFL~\cite{huang2023learning}, MEGE~\cite{pu2023memorizing}, LSTKC~\cite{xu2024lstkc}, CKP~\cite{xu2024mitigate} and DKP~\cite{xu2024distribution}). CIL is the most popular lifelong learning task and the CIL methods can be directly applied to the LReID task. 
Besides, the \textit{JointTrain} that indicates collecting the training data of all seen domains to train the model is also reported, which is commonly regarded as the upper bound of LReID methods~\cite{pu2023memorizing}. All experimental results are reported following the official publications or implemented with the official codes.

We present the performance of different methods on each seen domain, and the average performance across all seen domains (Seen-Avg) and unseen domains (UnSeen-Avg) in Tab. \ref{tab:setting1} and Tab. \ref{tab:setting2}, corresponding to Training Order-1 and Training Order-2 respectively. The best and second best results are highlighted in \textbf{Red} and \textbf{Blue}, separately.
\begin{figure}[t]
    \centering
	\includegraphics[width=1\linewidth]{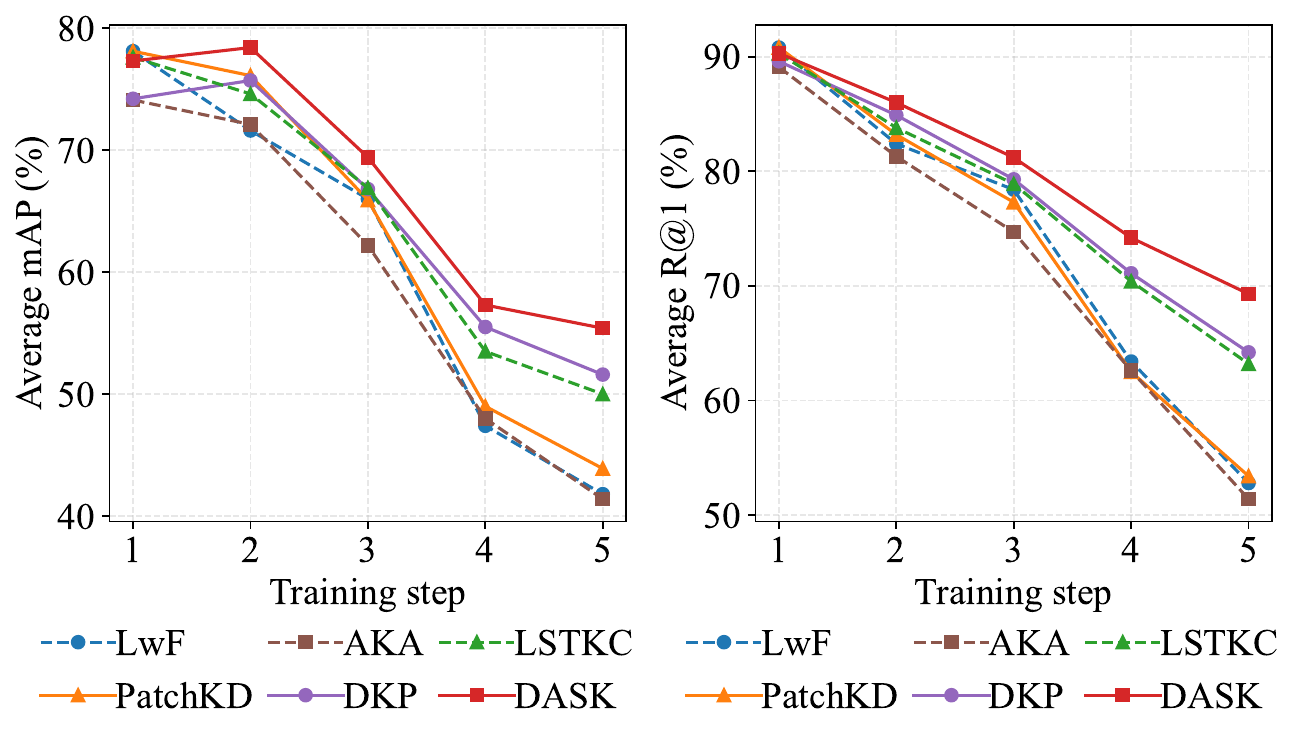}
        \caption{\label{fig:incremental} Anti-forgetting tendency on seen domains. }
\end{figure}

\begin{figure}[t]
    \centering
	\includegraphics[width=1\linewidth]{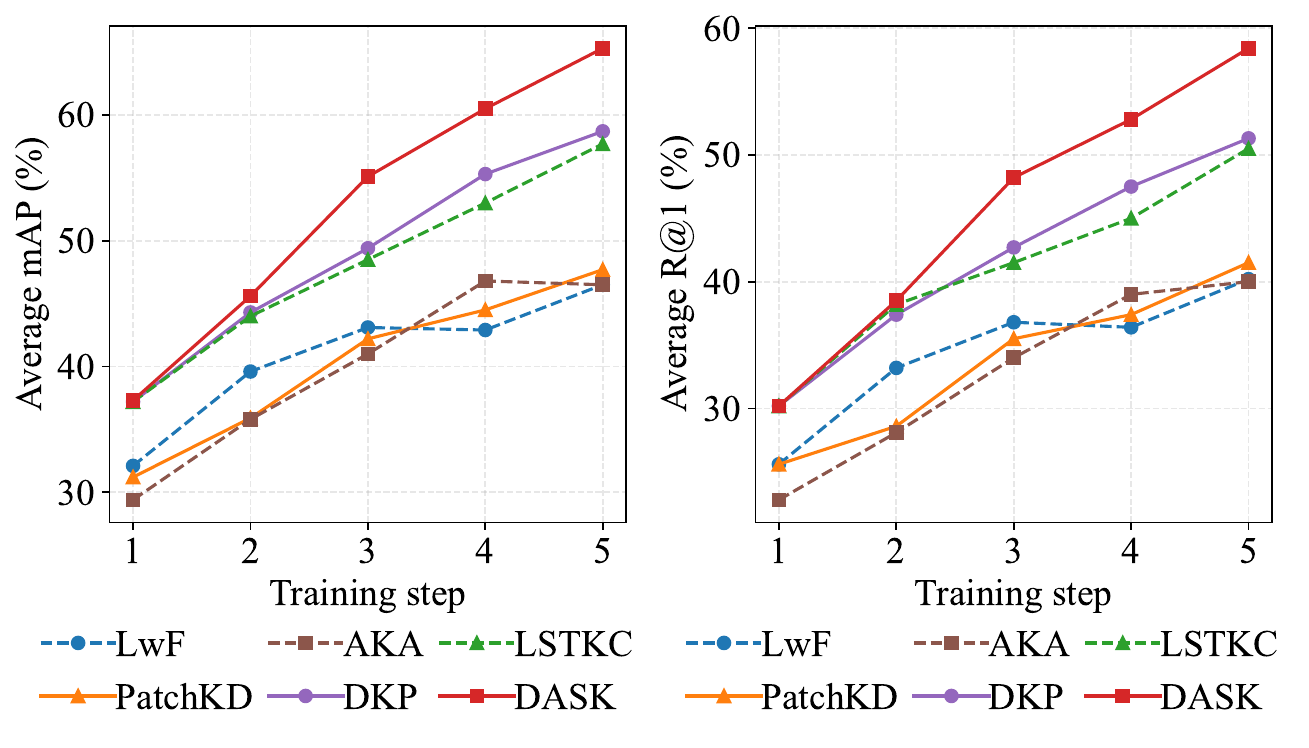}
        \caption{\label{fig:generalization}Generalization tendency on unseen domains.}
\end{figure}
\subsection{Seen-Domain Performance Evaluation}

\textbf{Compared to CIL Methods}: In Tab. \ref{tab:setting1} and Tab. \ref{tab:setting2}, our DASK outperforms all CIL methods on the first four subsets, obtaining \textbf{11.1\%/14.6\%} and \textbf{12.1\%/14.0\%} Seen-Avg mAP/R@1 improvements on two training orders. This is because CIL methods focus on improving the recognition capacity of known classes, whereas the identity classes of the test data are usually unseen during training. DASK obtains inferior results compared to PRAKA on the last subset CUHK03 due to the anti-forgetting designs of DASK partly limiting the learning of new data. The significantly superior average performance of DASK underscores that the proposed distribution rehearsing design can boost the model to achieve a better balance between learning and forgetting.

\textbf{Compared to LReID Methods}: 
In Tab. \ref{tab:setting1} and Tab. \ref{tab:setting2}, our DASK achieves leading performance on most subsets, with \textbf{3.6\%/5.2\%} and \textbf{5.8\%/6.8\%} Seen-Avg mAP/R@1 improvements compared to the state-of-the-art DKP. DASK achieves competitive results compared to PatchKD and ConRFL on the initial and last subsets, respectively. This is because PatchKD adopts strong anti-forgetting constraints and ConRFL adopts looser ones, resulting in over-emphasized performance on the first and last domains. In contrast, our method achieves a better balance between acquisition and forgetting under the proposed distribution rehearsing mechanism. As a result, significantly superior results on the middle subsets and Seen-Avg metrics are obtained.

\textbf{Seen Domain Performance Tendency}:
Fig. \ref{fig:incremental} shows the seen domain evaluation results across the training steps. Our DASK achieves similar performance with existing methods initially and consistently outperforms them from the second training step. These results arise because DASK adopts the same baseline as existing methods but the distribution rehearing design improves the knowledge consolidation capacity of our model during subsequent learning steps.

\subsection{Unseen-Domain Generalization Evaluation}
\textbf{Compared to CIL and LReID Methods}: 
As shown in the UnSeen-Avg terms in Tab. \ref{tab:setting1} and Tab.~\ref{tab:setting2}, our DASK outperforms all CIL and LReID methods by at least \textbf{6.1\%/6.1\%} and \textbf{6.5\%/4.5\%} mAP/R@1 on two training orders, which is a more significant improvement compared to the seen domains. This can be attributed to the distribution rehearsing design that enriches the data domains at each training step, guiding the model to learn both intra-domain and inter-domain discriminative knowledge, thus improving the model's generalization capacity to unknown conditions.

\textbf{UnSeen Domain Performance Tendency}: In Fig. \ref{fig:generalization}, our DASK gradually demonstrates significant superiority in unseen domain performance compared to existing methods as the training steps increase.
This is because our distribution rehearsing design enables the model to jointly learn from multiple seen domains to mine generalizable knowledge.

\begin{table}[t!]
  \centering
  \setlength{\tabcolsep}{1.6mm}{
    \begin{tabular}{lcccccccc}
    \hline
    & \multicolumn{2}{c}{Seen-Avg} & \multicolumn{2}{c}{UnSeen-Avg} \\
    \multirow{-1.9}[0]{*}{Method} &    mAP   & \multicolumn{1}{l}{R@1} & mAP   & \multicolumn{1}{l}{R@1} \\
          \hline
    Baseline                & 48.7 &60.7 &55.5 &48.2 \\
    Style Augmentation      & 48.6 & 63.1  & 59.9  & 52.3\\             
    Shared Convolution   &48.6&61.2&55.7&48.8 \\    
    Statistical Prediction (CoP)&51.2&65.1&61.3&	53.9	\\
    \hline
    \textbf{DASK}               & \textbf{55.4} & \textbf{69.3} & \textbf{65.3} & \textbf{58.4} \\
    \hline
    \end{tabular}%
 }
    \caption{Ablation on the input training data of ReID model.}    
  \label{tab:componment}%
\end{table}%

\begin{table}[t!]
  \centering
  \setlength{\tabcolsep}{1.5mm}{
    \begin{tabular}{ccccccccc}
    \hline
    &&& \multicolumn{2}{c}{Seen-Avg} & \multicolumn{2}{c}{UnSeen-Avg} \\
    \multirow{-1.9}[0]{*}{Baseline} &\multirow{-1.9}[0]{*}{$\mathcal{L}_{ReID}^*$}&\multirow{-1.9}[0]{*}{{$\mathcal{L}_{SKD}^*$}}&    mAP   & \multicolumn{1}{l}{R@1} & mAP   & \multicolumn{1}{l}{R@1} \\
          \hline
    $\checkmark$& \ding{55}&  \ding{55} & 48.7 &60.7 &55.5 &48.2 \\
    $\checkmark$&$\checkmark$ & \ding{55}&55.0&    68.8&    63.4&    56.0     \\             
    $\checkmark$& \ding{55} &$\checkmark$& 49.9&    61.5&    56.3&    49.2  \\    
    $\checkmark$&$\checkmark$&$\checkmark$&\textbf{55.4} & \textbf{69.3} & \textbf{65.3} & \textbf{58.4} \\
    \hline
    \end{tabular}%
 }
    \caption{Ablation on the rehearsing-guided loss components. }    
  \label{tab:loss}%
\end{table}%

\subsection{Ablation Studies}
\label{sec:ablation}
We conduct ablation studies on the proposed components under Training Order-1 of the LReID benchmark as follows.

\textbf{Ablation on Distribution Rehearsing Strategy}: \label{sec:ablation-distribution}
In Tab.~\ref{tab:componment}, additional distribution rehearsing choices are compared with our DASK: (1) Baseline denotes using only the new data for training. (2) Style Augmentation denotes adopting the distribution-augmented data in Fig.~\ref{fig:framework} (b). (3) Shared Convolution denotes using a shared convolution kernel for all images. (4) Statistical Prediction indicates predicting only the mean and standard deviation of the transferred image, as conducted by CoP~\cite{gu2023color}. The results show our DASK significantly outperforms the other solutions. Note that Style Augmentation can generate unreasonably hard samples that disturb model learning, Shared Convolution applies the same transformation to all images, neglecting instance-level distribution alignment, and Statistical Prediction ignores informative contextual clues. As a result, the distribution rehearsing strategy of DASK generates higher-quality old-style data and improves lifelong learning due to the adaptive distribution transfer kernel design.

\textbf{Effectiveness of Rehearsing-guided Loss Components}: 
In Tab.~\ref{tab:loss}, Baseline denotes training with $\mathcal{L}_{ReID}$ and $\mathcal{L}_{SKD}$. By progressively incorporating the $\mathcal{L}_{ReID}^*$ and $\mathcal{L}_{SKD}^*$, the results demonstrate marked improvements in LReID performance. Specifically, $\mathcal{L}_{ReID}^*$ facilitates the network in reinforcing discriminative features from old domains. Concurrently, $\mathcal{L}_{SKD}^*$ enhances structural consistency, which mitigates the forgetting caused by parameter overwriting. 
\begin{figure}[t]
    \centering
	\includegraphics[width=0.95\linewidth]{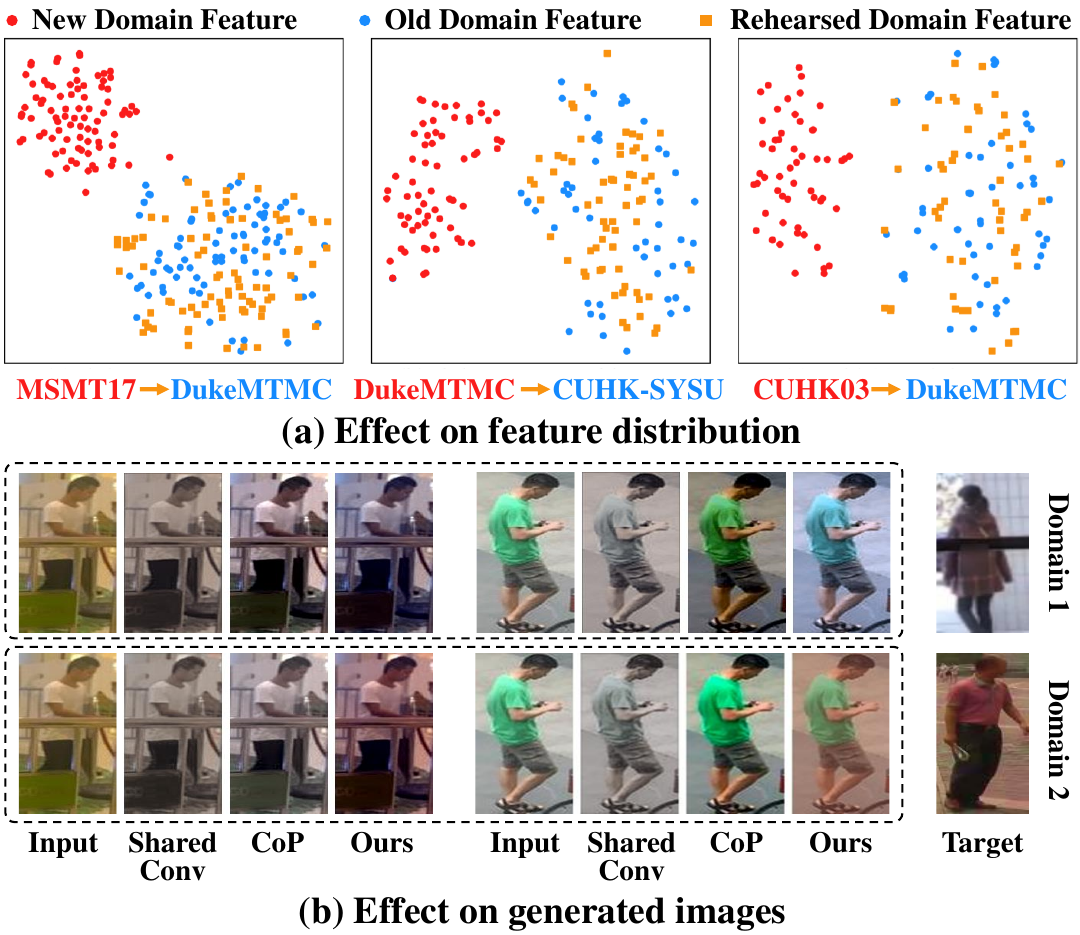}
        \caption{\label{fig:feature-distribution} Visualization of distribution rehearsing effects.}
\end{figure}

\textbf{Visualization results}: 
Fig. \ref{fig:feature-distribution} (a) is the t-SNE visualization of the features of the new, old, and rehearsed (transferred from new) domains. There are obvious gaps between new and old feature distributions, while the distributions of rehearsed features and old features overlap significantly.
These results demonstrate that our domain rehearsing strategy can effectively generate old-style data. 
Besides, Fig.~\ref{fig:feature-distribution} (b) visualizes generated images of Shared Convolution, CoP and ours. It is evident that our generations more closely resemble the styles of various target domains, supporting the analysis in Sec. \textit{Ablation on Distribution Rehearsing Strategy}.

\section{Conclusion}
In this paper, we propose DASK, a novel exemplar-free LReID method. The key idea is a paradigm that models and rehearses the distribution of the old domains to enhance lifelong knowledge consolidation. To achieve this, a Distribution Rehearser Learning mechanism is introduced, where an adaptive kernel prediction network (AKPNet) is designed to generate an instance-specific domain transfer kernel, achieving a fine-grained style adjustment. Besides, a Distribution Rehearsing-driven LReID Training scheme is introduced to accomplish new knowledge learning and old knowledge accumulation.
Our results underscore that the input distribution is closely relevant to the learned knowledge in LReID, and the image level distribution rehearsing is effective for addressing catastrophic forgetting. In future work, we will investigate extending our distribution rehearing design to more lifelong learning tasks.

\section{Acknowledgments}
This work was supported by grants from the National Natural Science Foundation of China (62376011, 61925201, 62132001, 62432001) and Beijing Natural Science Foundation (L247006).
\appendix

\section{Appendix}
In this appendix, we provide supplementary experimental results, analysis, and implementation details of our proposed method. The content is organized as follows:
\begin{itemize}
\item \textbf{Comparison with State-of-the-Art Replay-Based LReID Methods}: We present a comprehensive evaluation of the proposed DASK against state-of-the-art Replay-based LReID approaches to highlight its effectiveness.
\item \textbf{Ablation Studies on Hyperparameters and Model Components}: We investigate the influence of key hyperparameters and model components on the performance of DASK, providing insights into the design choices.
\item \textbf{LReID Benchmark Datasets}: We offer a detailed overview of the datasets used in the LReID benchmark.
\end{itemize}

\begin{table*}[htbp]
   \centering
  \setlength{\tabcolsep}{0.9mm}{
    \begin{tabular}{llccccccccccc>{\columncolor{seen_back}}c>{\columncolor{seen_back}}c>{\columncolor{unseen_back}}c>{\columncolor{unseen_back}}cccc}   
    \hline
   &\multirow{2}[1]{*}{Method}&\multirow{2}[1]{*}{Publication} & \multicolumn{2}{c}{Market-1501} & \multicolumn{2}{c}{CUHK-SYSU} & \multicolumn{2}{c}{DukeMTMC} & \multicolumn{2}{c}{MSMT17} & \multicolumn{2}{c}{CUHK03} & \multicolumn{2}{>{\columncolor{seen_back}}c}{\textbf{Seen-Avg}} & \multicolumn{2}{>{\columncolor{unseen_back}}c}{\textbf{UnSeen-Avg}} \\
\hhline{~~~--------------}
&& & mAP   & R@1   & mAP   & R@1   & mAP   & R@1   & mAP   & R@1   & mAP   & R@1   & mAP   & R@1   & mAP   & R@1 \\
\hline   
&JointTrain   &-  &78.9  & 90.9  & 86.7  & 88.2  & 71.2  & 82.9  & 36.2  & 61.2  & 61.2  & 63.4  & 66.8  & 77.3  & 59.4  & 52.6  \\  
    \hline
\multirow{4}[0]{*}{\rotatebox{90}{Replay} }   
&iCaRL      &ICCV 2017   &  52.1  & 74.5  & 81.3  & 83.6  & 48.0  & 65.6  & 18.9  & 42.5  & 44.1  & 45.4  & 48.9  & 62.3  & 51.9  & 45.9  \\
&GwFReID    & AAAI 2021  &57.7  & 77.4  & 79.6  & 81.7  & 48.5  & 66.4  & 22.0  & 45.0  &\textcolor{orange}{\textbf{58.8}}  &\textcolor{orange}{\textbf{61.4}}  & 53.3  & 66.4  & 51.4  & 44.9  \\
&PTKP       & AAAI 2022  &\textcolor{orange}{\textbf{73.5}}  &\textcolor{orange}{\textbf{88.0}}  &\textcolor{orange}{\textbf{84.8}}  &\textcolor{orange}{\textbf{86.6}}  &\textcolor{orange}{\textbf{59.0}}  & 75.0  & 23.2  & 47.1  &51.6  &53.8 &\textcolor{orange}{\textbf{58.4}}  &\textcolor{orange}{\textbf{70.1}}  & 57.5  & 51.1  \\
&KRKC       & AAAI 2023  & 60.2  & 83.6  & 84.1  & 86.3  & 58.9  &\textcolor{orange}{\textbf{76.6}}  &\textcolor{orange}{\textbf{24.2}}  &\textcolor{orange}{\textbf{51.5}}  & 43.1  & 44.3  & 54.1  & 68.5  &\textcolor{orange}{\textbf{59.4}}  &\textcolor{orange}{\textbf{53.0}}  \\

        \hline
\multirow{5}[0]{*}{\rotatebox{90}{No-Replay} }         
&iCaRL\dag      &ICCV 2017   &36.1  &62.9  &76.8  & 79.5  & 34.0  & 54.4  &12.6  & 32.4  & 46.7  & 47.8  & 41.2  &55.4  &52.3 &46.2  \\
&GwFReID\dag    & AAAI 2021  &32.5  & 57.8  & 69.8  & 72.7  & 26.7  & 43.9  & 8.6   & 23.4  &\textcolor{red}{\textbf{60.5}}  &\textcolor{red}{\textbf{63.4}}  & 39.6  & 52.2  & 48.5  & 42.1  \\
&PTKP\dag       & AAAI 2022  &\textcolor{blue}{\textbf{58.0}}  &\textcolor{blue}{\textbf{79.8}}  &\textcolor{red}{\textbf{82.7}}  &\textcolor{red}{\textbf{84.8}}  &\textcolor{blue}{\textbf{47.1}}  &\textcolor{blue}{\textbf{64.9}}  &\textcolor{blue}{\textbf{17.4}}  &\textcolor{blue}{\textbf{39.1}}  &\textcolor{blue}{\textbf{48.1}}  &\textcolor{blue}{\textbf{49.4}}  &\textcolor{blue}{\textbf{50.7}}  &\textcolor{blue}{\textbf{63.6}}  &\textcolor{blue}{\textbf{58.8}}  &\textcolor{blue}{\textbf{51.9}}  \\
&KRKC\dag       & AAAI 2023  &33.3  & 62.7  & 76.8  & 80.6  & 38.2  & 59.2  & 12.1  & 32.6  & 22.5  & 22.1  & 36.6  & 51.4  & 48.2  & 43.3  \\

    \hhline{~----------------}
&\textbf{DASK} & This Paper&\textcolor{red}{\textbf{61.2}}  &\textcolor{red}{\textbf{82.3}}  &\textcolor{blue}{\textbf{81.9}}  &\textcolor{blue}{\textbf{83.7}}  &\textcolor{red}{\textbf{58.5}}  &\textcolor{red}{\textbf{74.6}}  &\textcolor{red}{\textbf{29.1}} &\textcolor{red}{\textbf{ 57.6}}  & 46.2  & 48.1  &\textcolor{red}{\textbf{55.4}}  & \textcolor{red}{\textbf{69.3}}  &\textcolor{red}{\textbf{65.3}}  &\textcolor{red}{\textbf{58.4}}  \\   
\hline
    \end{tabular}%
   }    
  \caption{
Comparison results of Seen-domain anti-forgetting and UnSeen-domain generalization on Training Order-1. 
$\dag$ indicates the results re-implemented without replaying historical exemplars.
  }  
  \label{tab:setting1}%
\end{table*}%

\begin{table*}[htbp]
  \centering
  \setlength{\tabcolsep}{0.9mm}{
    \begin{tabular}{llccccccccccc>{\columncolor{seen_back}}c>{\columncolor{seen_back}}c>{\columncolor{unseen_back}}c>{\columncolor{unseen_back}}cccc}
    \hline
    &\multirow{2}[1]{*}{Method}&\multirow{2}[1]{*}{Publication}& \multicolumn{2}{c}{DukeMTMC} & \multicolumn{2}{c}{MSMT17} & \multicolumn{2}{c}{Market-1501} & \multicolumn{2}{c}{CUHK-SYSU} & \multicolumn{2}{c}{CUHK03} & \multicolumn{2}{>{\columncolor{seen_back}}c}{\textbf{Seen-Avg}} & \multicolumn{2}{>{\columncolor{unseen_back}}c}{\textbf{UnSeen-Avg}} \\
\hhline{~~~--------------}
&& & mAP   & R@1   & mAP   & R@1   & mAP   & R@1   & mAP   & R@1   & mAP   & R@1   & mAP   & R@1   & mAP   & R@1 \\
\hline
&JointTrain &-& 71.2  & 82.9  & 36.2  & 61.2  & 78.9  & 90.9  & 86.7  & 88.2  & 61.2  & 63.4  & 66.8  & 77.3  & 59.4  & 52.6  \\
\hline   
\multirow{4}[0]{*}{\rotatebox{90}{Replay} }   
&iCaRL      &ICCV 2017      &45.4  & 64.3  & 16.7  & 37.9  & 62.4  & 83.1  & 83.4  & 85.4  & 44.9  & 46.8  & 50.5  & 63.5  & 56.2  & 50.3  \\
&GwFReID    & AAAI 2021  &46.4  & 64.1  & 16.5  & 36.3  & 65.7  & 83.7  & 80.4  & 82.4  &\textcolor{orange}{\textbf{58.4}}  &\textcolor{orange}{\textbf{61.5}}  & 53.5  & 65.6  & 52.2  & 45.9  \\
&PTKP       & AAAI 2022  &\textcolor{orange}{\textbf{58.6}}  &\textcolor{orange}{\textbf{74.3}}  & 16.4  & 37.3  & 67.1  & 84.8  & 83.1  & 85.1  & 49.8  & 52.9  &\textcolor{orange}{\textbf{55.0}}  &\textcolor{orange}{\textbf{66.9}}  & 57.5  & 51.1  \\
&KRKC       & AAAI 2023  & 50.1  & 68.6  &\textcolor{orange}{\textbf{17.7}}  &\textcolor{orange}{\textbf{41.1}}  &\textcolor{orange}{\textbf{69.0}}  &\textcolor{orange}{\textbf{88.3}}  &\textcolor{orange}{\textbf{85.2}}  &\textcolor{orange}{\textbf{87.4}}  & 40.4  & 41.6  & 52.5  & 65.4  &\textcolor{orange}{\textbf{59.4}}  &\textcolor{orange}{\textbf{53.4}}  \\

        \hline
\multirow{5}[0]{*}{\rotatebox{90}{No-Replay} }         
&iCaRL\dag      &ICCV 2017      &28.1  & 45.4  & 8.5   & 23.1  & 48.9  & 73.6  & 78.9  & 81.2  & 49.2  & 51.5  & 42.7  & 55.0  & 54.1  & 47.3  \\
&GwFReID\dag    & AAAI 2021  &22.1  & 37.8  & 4.7   & 14.0  & 46.7  & 68.9  & 72.4  & 75.7  &\textcolor{red}{\textbf{62.2}}  &\textcolor{red}{\textbf{64.7}}  & 41.6  & 52.2  & 48.6  & 41.4  \\
&PTKP\dag       & AAAI 2022  &\textcolor{blue}{\textbf{50.0}}  &\textcolor{blue}{\textbf{67.4}}  &\textcolor{blue}{\textbf{13.1}}  &\textcolor{blue}{\textbf{31.3}}  &\textcolor{blue}{\textbf{56.7}}  &\textcolor{blue}{\textbf{77.8}}  &\textcolor{blue}{\textbf{83.3}}  &\textcolor{blue}{\textbf{85.4}}  &\textcolor{blue}{\textbf{50.2}}  &\textcolor{blue}{\textbf{51.9}}  &\textcolor{blue}{\textbf{50.7}}  &\textcolor{blue}{\textbf{62.8}}  &\textcolor{blue}{\textbf{57.0}}  &\textcolor{blue}{\textbf{49.9}} \\
&KRKC\dag       & AAAI 2023  &  30.2  & 49.9  & 8.7   & 25.5  & 45.8  & 74.7  & 78.0  & 82.0  & 21.5  & 21.0  & 36.8  & 50.6  & 46.7  & 41.9  \\

    \hhline{~----------------}
&\textbf{DASK} & This Paper&\textcolor{red}{\textbf{55.7}}  &\textcolor{red}{\textbf{74.4}}  & \textcolor{red}{\textbf{25.2}}  & \textcolor{red}{\textbf{51.9}}  & \textcolor{red}{\textbf{71.6}}  & \textcolor{red}{\textbf{87.7}}  & \textcolor{red}{\textbf{84.8}}  & \textcolor{red}{\textbf{86.2}}  & 48.4  & 49.8  & \textcolor{red}{\textbf{57.1}}  & \textcolor{red}{\textbf{70.0}}  & \textcolor{red}{\textbf{65.5}}  & \textcolor{red}{\textbf{57.9}}  \\ 
\hline
    \end{tabular}%
   }   
  \caption{
Comparison results of Seen-domain anti-forgetting and UnSeen-domain generalization on Training Order-2. 
$\dag$ indicates the results re-implemented without replaying historical exemplars.
  }  
  \label{tab:setting2}%
\end{table*}%

\begin{figure*}[htbp]
\centering
\subfloat[\label{fig:lambda-1} Loss weight $\alpha$]{
    \includegraphics[scale=0.45]{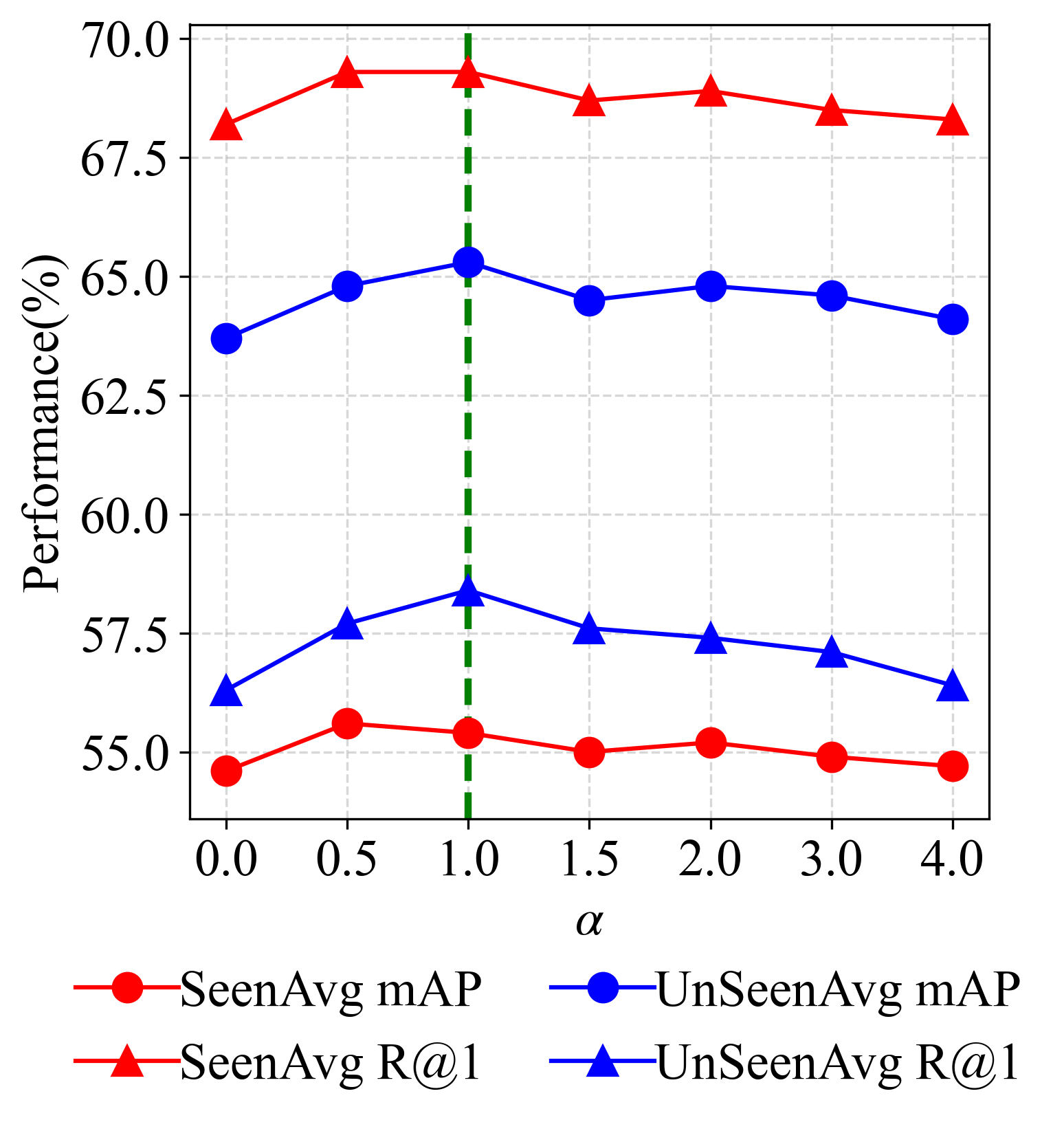}}
\subfloat[\label{fig:lambda-2}  Loss weight $\beta$]{
    \includegraphics[scale=0.45]{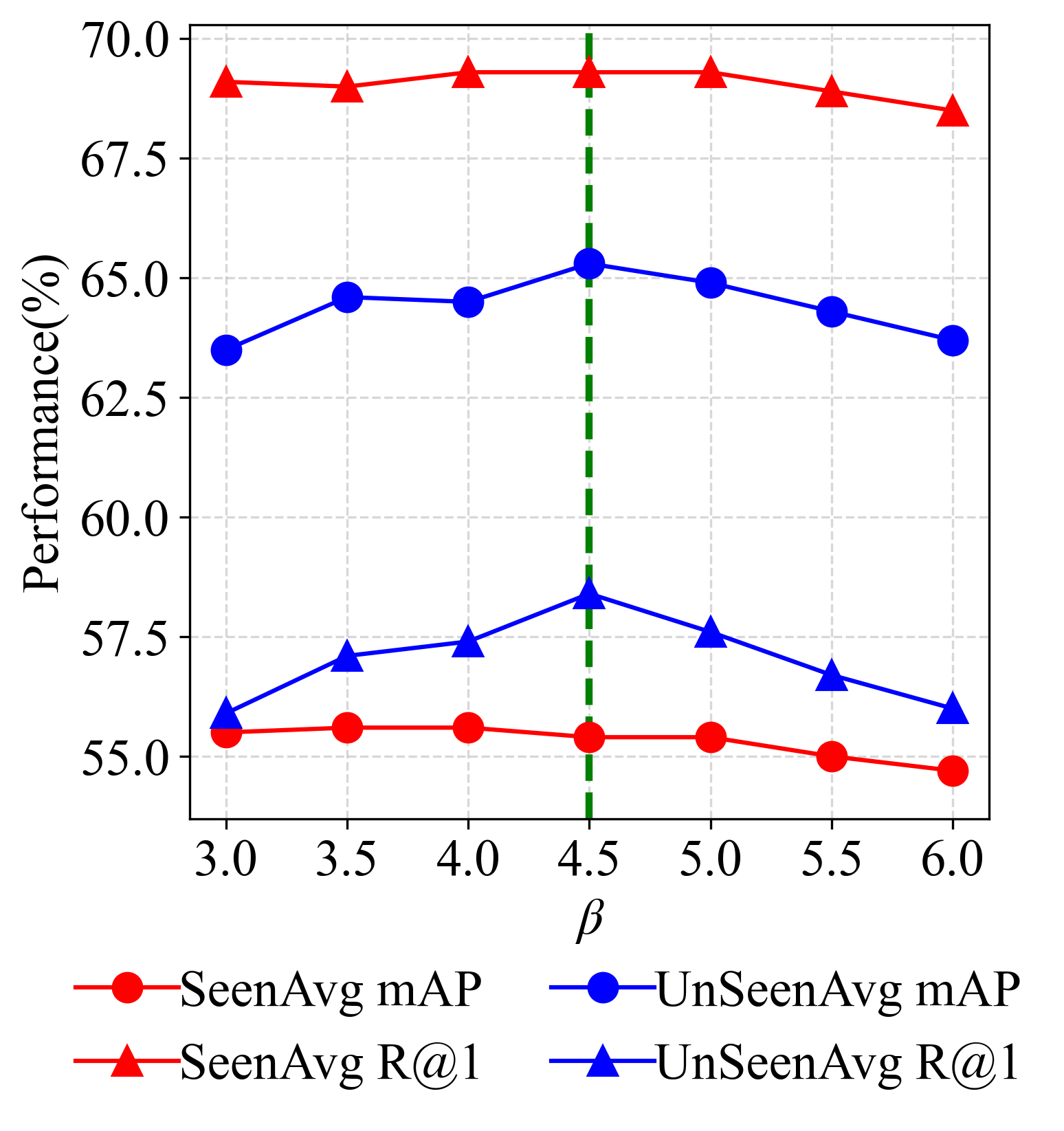}}
\subfloat[\label{fig:proto-sample} Old model EMA weight $\lambda$]{
    \includegraphics[scale=0.45]{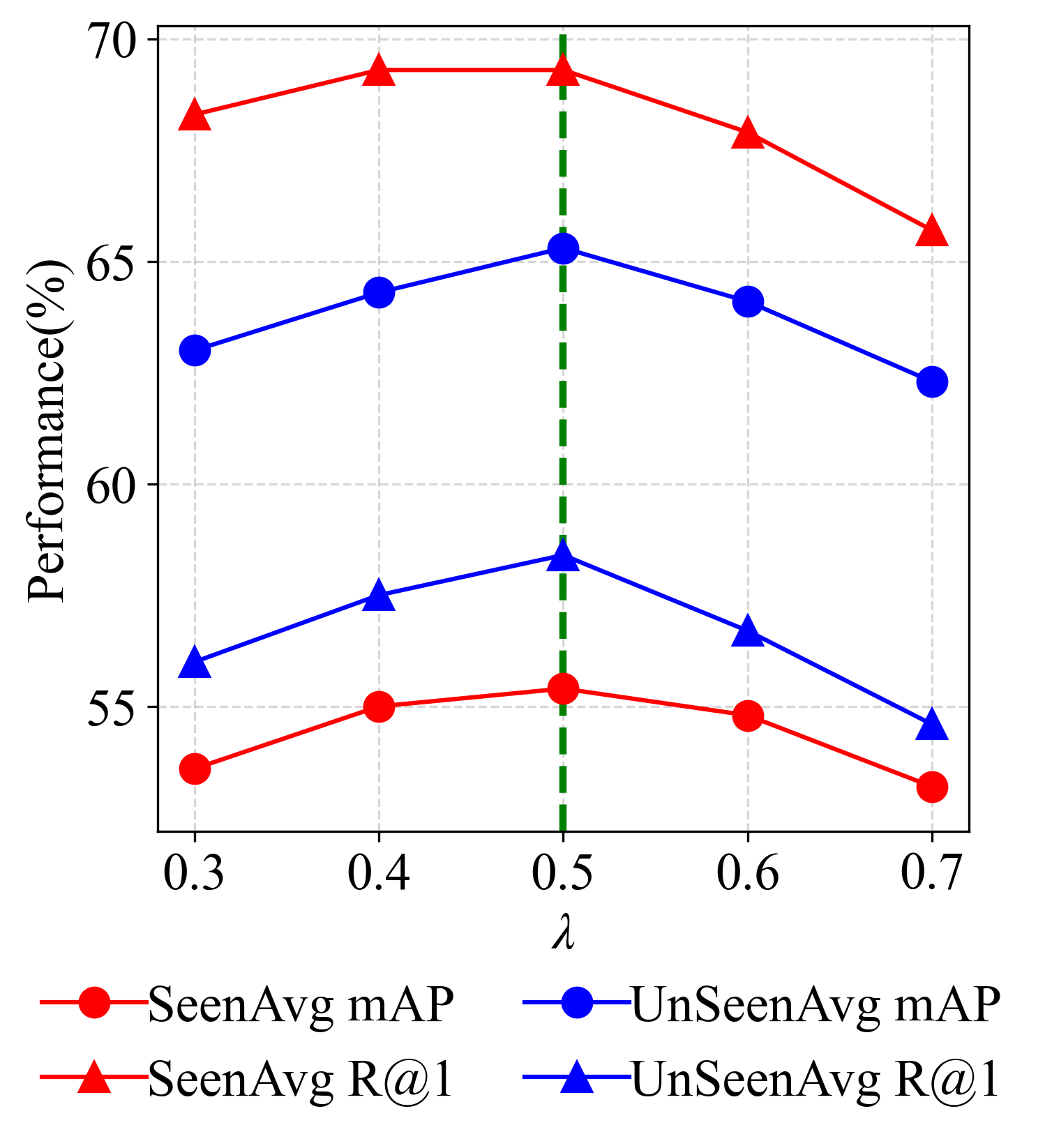}}

\caption{Ablation studies on the hyperparameters in this paper. 
       The values marked by the dashed lines are adopted by our proposed method.}
\label{fig:ablation_supple} 
\end{figure*}

\section{Comparsion with Replay-based Methods}
 
\subsection{Compared Methods}
 We compare with existing \textbf{Reply-based} LReID methods, including GeFReID~\cite{wu2021generalising}, PTKP~\cite{ge2022lifelong}, and KRKC~\cite{yu2023lifelong}. Additionally, the results of the classical replay-based CIL method, iCaRL~\cite{rebuffi2017icarl} are also reported. Besides, for a fair comparison with our exemplar-free approach, DASK, we also present the \textbf{No-Replay} results for these methods by removing the data replay designs. The corresponding methods are noted as GeFReID$\dag$, PTKP$\dag$, KRKC$\dag$, and iCaRL$\dag$, respectively. 
The results on each seen domain, as well as the average performance across all seen domains (Seen-Avg) and unseen domains (UnSeen-Avg), are presented in Tab. \ref{tab:setting1} and Tab. \ref{tab:setting2}, corresponding to Training Order-1 and Training Order-2 respectively. All experiments are conducted using the official codes or implemented following the original papers. Note that the best  \textbf{Replay-based} results are marked as \textcolor{orange}{\textbf{Orange}}, and the best and second best \textbf{No-Replay} setting results are separately highlighted in \textcolor{red}{\textbf{Red}} and \textcolor{blue}{\textbf{Blue}}.

\textbf{Replay Results \textit{vs.} DASK}: As shown in Tab. \ref{tab:setting1} and Tab. \ref{tab:setting2}, our DASK achieves competitive results compared to the state-of-the-art Replay-based methods on each subsets. Although PTKP outperforms DASK by 3.0\%/0.8\% in Seen-Avg mAP/R@1 on the training order-1 (Tab. \ref{tab:setting1}), our DASK surpasses all Replay-based methods on training order-2 (Tab. \ref{tab:setting2}), with \textbf{2.1\%/0.1\%} improvement in Seen-Avg mAP/R@1. Additionally, DASK demonstrates significantly superior generalization capacity on unseen domains compared to all existing Replay-based methods, with a \textbf{5.9\%/5.4\%} and \textbf{6.1\%/4.5\%} improvement in UnSeen-Avg mAP/R@1 on Training Order-1 and Training Order-2, respectively.

The competitive and superior results on seen domains compared to state-of-the-art Replay-based methods are attributed to the distribution rehearsing mechanism, which enables our method to generate abundant old-style data, largely mitigating the catastrophic forgetting problem caused by the absence of historical patterns. Furthermore, the significant improvement in unseen domains is due to the fact that, in Replay-based methods, the identities across different domains do not overlap, causing the model to take the image style differences between domains as discriminative clues, thereby limiting its generalization to unseen domains, especially those with high intra-domain diversity. In contrast, the APKNet of DASK can generate diverse-style images of the same identity, guiding the model to extract domain-relevant discriminative features, which enables DASK to generalize better under different unseen conditions.

 \begin{figure*}[htbp]
		\begin{center}
			\includegraphics[width=0.9\linewidth]{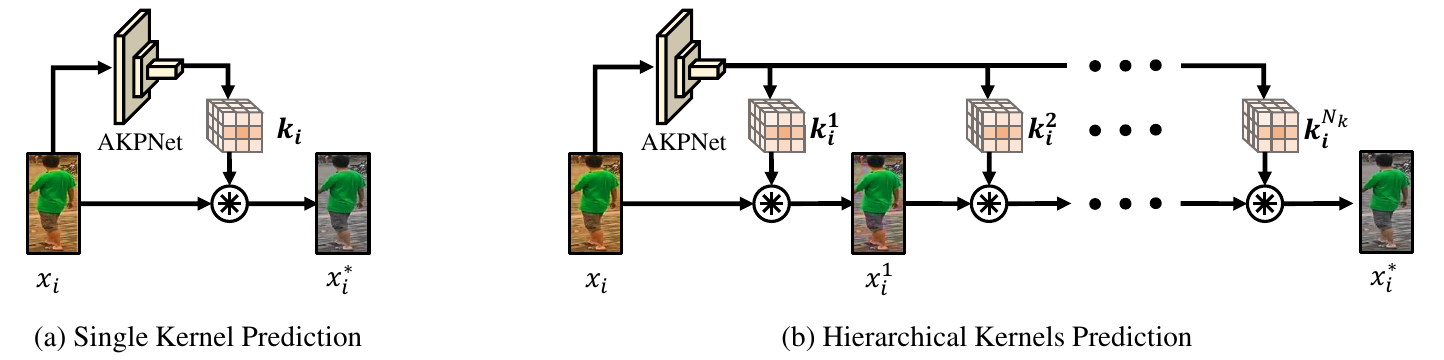}
			\caption{Variants of predictable style transfer kernel designs. (a) Predicting a single-layer style transfer kernel. (b) Predicting multiple ($N_k$) layer style transfer kernels to achieve hierarchical generation.}
			\label{fig:multi-kernel}
		\end{center}    
	\end{figure*}

\begin{figure}[th]
    \begin{center}
        \includegraphics[width=1.0\linewidth]{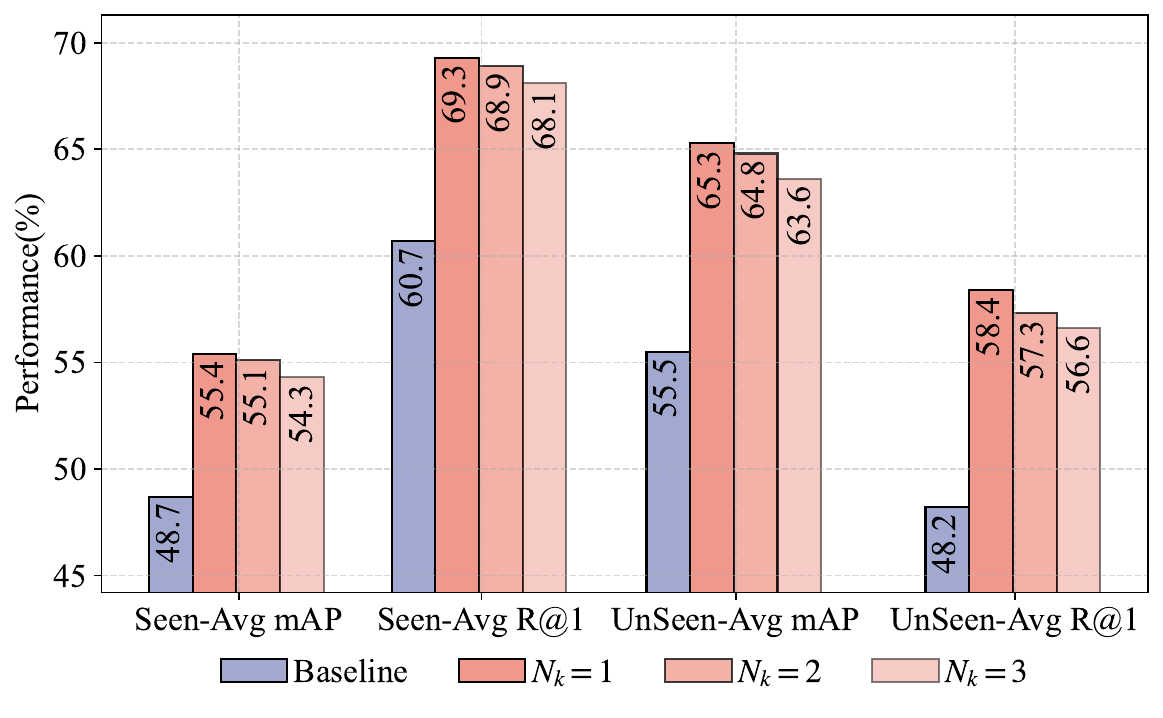}
        \caption{Results comparison of Baseline and DASK with different style transfer kernel layer numbers $N_k$.}
        \label{fig:multi-kernel-res}
    \end{center}    
\end{figure}

\textbf{No-Replay Results \textit{vs.} DASK}:
When exemplar replay is prohibited, all Replay-based methods exhibit reduced anti-forgetting capacity, while our DASK consistently achieves superior performance across most subsets. Specifically, DASK delivers a \textbf{4.7\%/5.7\%} and \textbf{6.4\%/7.2\%} improvement in Seen-Avg mAP/R@1, as well as a \textbf{6.5\%/6.5\%} and \textbf{8.5\%/8.0\%} enhancement in UnSeen-Avg mAP/R@1 across two training orders in Tab. \ref{tab:setting1} and Tab. \ref{tab:setting2}. These results highlight the effectiveness of our DASK method in consolidating knowledge during lifelong learning. 
We also observe that GwFReID$\dag$ and PTKP$\dag$ achieve higher performance on the CUHK3 dataset. 
This can be attributed to these methods primarily relying on data replay to mitigate forgetting, which has a smaller impact on new data learning compared to our DASK. However, this advantage is based on their limited anti-forgetting capacity, thus our DASK achieves significantly superior average results across different domains.

\section{Additional Ablation Studies}
\subsection{Influence of Hyperparameters}
 In our paper, $\alpha$, $\beta$, and $\lambda$ in Eq. (5) and Eq. (6) are primary hyperparameters. 
 In this section, we conduct experiments to explore the impact of different hyperparameter values on model performance.  
 When adjusting one hyperparameter, all other hyperparameters are set to their default values.
 
 As shown in \ref{fig:ablation_supple} (a), the model performance rises initially and then decreases as the relation distillation loss weight $\alpha$ increases. This is because $\mathcal{L}_{SKD}$ and $\mathcal{L}_{SKD}^*$ help reduce the forgetting of historical knowledge. However, when the loss weight $\alpha$ is too large, the model's plasticity is diminished, limiting its performance on new data. Thus, a proper $\alpha$  value can lead to a good balance between knowledge acquisition and forgetting.
 
 As shown in  \ref{fig:ablation_supple} (b), the seen domain performance is stable as the rehearsing-guided loss weight $\beta$ changes, while the unseen domain performance is more sensitive to $\beta$ values. This is because the presence or absence of historical patterns primarily affects the model's anti-forgetting capacity, whereas the weights of different loss components are less influential. In contrast, since distribution rehearsing introduces diverse-style data simultaneously, balancing the impact of different styles is crucial for the model to capture generalizable knowledge.

 In \ref{fig:ablation_supple} (c), we visualize the influence of EMA weight $\lambda$ on the model performance. It is evident that both too small and too large $\lambda$ values lead to degraded performance. This occurs because, in these conditions, the new and old knowledge is over-emphasized, respectively.

 Therefore, in this paper, we set $\alpha$, $\beta$, and $\lambda$ to 1.0, 4.5, and 0.5 by default.

\subsection{Influence of Predicted Kernel Number}
In our original paper, we are inspired by the existing domain generalization works, which reveal that the data styles are encoded in the low-level statistics of images \cite{ulyanov2016texture, hong2021stylemix}. To meet the statistics transfer assumption and adapt to the potential texture changes, we use a predictable convolution kernel to accomplish distribution transfer (as shown in Fig. ~\ref{fig:multi-kernel} (a)).

In this supplementary, we further explore accomplishing style transfer via a multi-layer design, \textit{i.e.}, generating multiple style transfer kernels to achieve target image generation hierarchically (as shown in Fig. ~\ref{fig:multi-kernel} (b)). Specifically, given an input image $x_i$ a group of kernels $\mathcal{K}=\{\boldsymbol{k}_i^j\in\mathbb{R}^{C\times C\times k \times k}\}_{j=1}^{N_k}$ is predicted by the AKPNet, where $N_k$ is the number of kernels. Therefore, the style transfer process can be presented as $x_i^*=\boldsymbol{k}_i^{N_k}\circledast(...(\boldsymbol{k}_i^1\circledast x_i))$. Note that Fig. ~\ref{fig:multi-kernel} (a) is a special case of Fig. ~\ref{fig:multi-kernel} (b) when $N_k=1$.

Fig. ~\ref{fig:multi-kernel-res} visualizes the results of the Baseline and different values of $N_k$. The results indicate that the model achieves optimal performance when $N_k=1$, while larger values of $N_k$ lead to degraded performance. This degradation occurs because multi-layer style transfer introduces too many parameters, making the model prone to overfitting the seen textures. 
\begin{figure}[t]
    \begin{center}
        \includegraphics[width=1.0\linewidth]{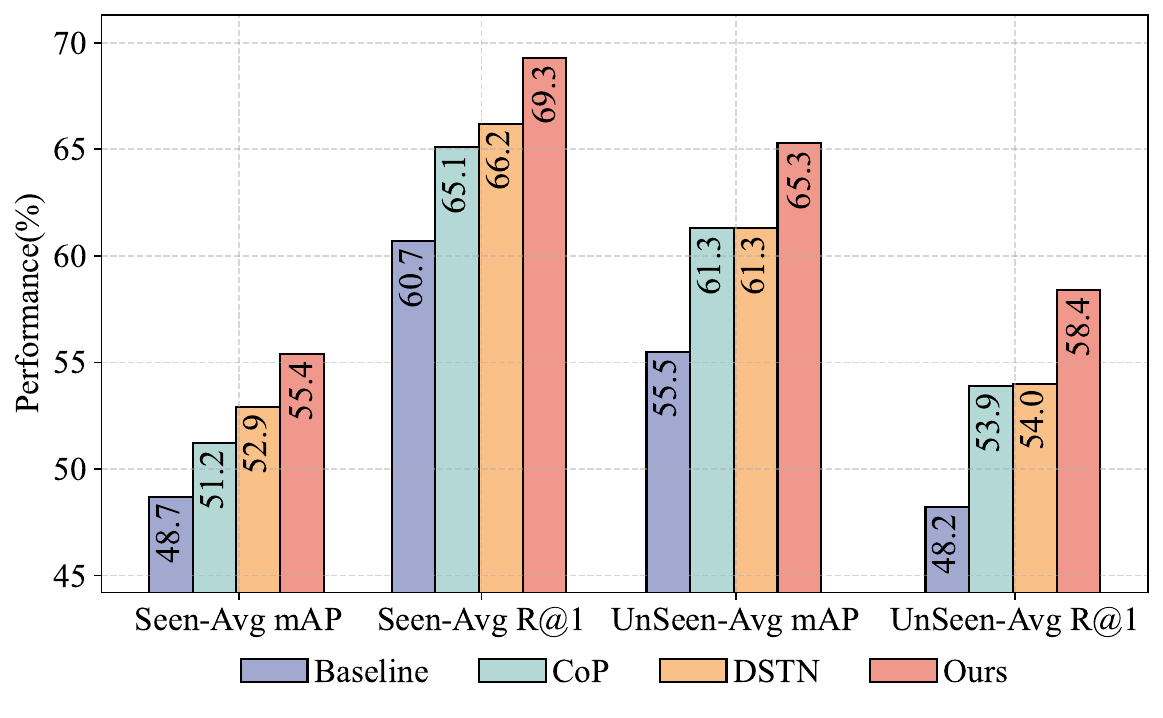}
        \caption{Results comparison under different style transfer approaches.}
        \label{fig:network}
    \end{center}    
\end{figure}

 \begin{table*}[t]
  \centering
  
  \setlength{\tabcolsep}{3.4mm}{
    \begin{tabular}{l|l|llc|ccc}
    \hline
    \multirow{2}[0]{*}{Type} & \multirow{2}[0]{*}{Datasets Name} & \multicolumn{3}{c|}{Original Identities} & \multicolumn{3}{c}{LReID Identities} \\
\cline{3-8}          &       & Train & Query & Gallery & Train & Query & Gallery \\
    \hline
    \multirow{5}[2]{*}{Seen} & CUHK03~\cite{li2014deepreid} & 767   & 700   & 700   & 500   & 700   & 700 \\
          & Market-1501~\cite{zheng2015scalable} & 751   & 750   & 751   & 500   & 751   & 751 \\
          & DukeMTMC-ReID~\cite{ristani2016performance} & 702   & 702   & 1110  & 500   & 702   & 1110 \\
          & CUHK-SYSU~\cite{xiao2016end} & 942   & 2900  & 2900  & 500   & 2900  & 2900 \\
          & MSMT17-V2~\cite{wei2018person}& 1041  & 3060  & 3060  & 500   & 3060  & 3060 \\          
    \hline
    \multirow{7}[2]{*}{Unseen}
    & i-LIDS~\cite{branch2006imagery}& 243   & 60    & 60    & -      & 60    & 60 \\
    & VIPR~\cite{gray2008viewpoint} & 316   & 316   & 316   &  -     & 316   & 316 \\
    & GRID~\cite{loy2010time} & 125   & 125   & 126   &  -     & 125   & 126 \\   
          & PRID~\cite{hirzer2011person}& 100   & 100   & 649   & -      & 100   & 649 \\
                 
          & CUHK01~\cite{li2012human} & 485   & 486   & 486   & -      & 486   & 486 \\
          & CUHK02~\cite{li2013locally} & 1677  & 239   & 239   & -      & 239   & 239 \\
          & SenseReID~\cite{zhao2017spindle}& 1718  & 521   & 1718  & -      & 521   & 1718 \\
    \hline
    \end{tabular}%
    }
    \caption{The statistics of datasets used in the LReID benchmark. ‘-’ indicates the datasets are only used for testing.}
  \label{tab:datasets}%
\end{table*}%

\begin{figure}[t]
    \begin{center}        \includegraphics[width=1.0\linewidth]{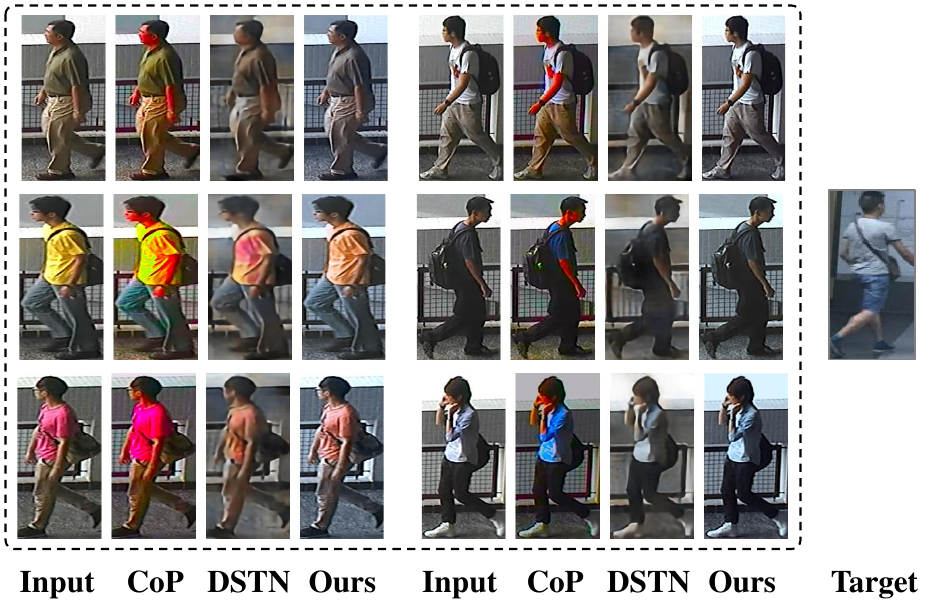}
        \caption{Generation visualization of different style transfer approaches.}
        \label{fig:vgg-reconstruction}
    \end{center}    
\end{figure}

\subsection{Comparison with Deep Generation Network}
Previous studies~\cite{hong2021domain} have utilized VGG-based encoder-decoder networks to perform style transfer. In this supplementary material, we adopt the same network architecture as DSTN \cite{hong2021domain} to train an end-to-end generation network. This network is then employed as a substitute for the DRRT module (the style transfer component in our AKPNet) within our DASK framework.

Fig~\ref{fig:network}  illustrates the results of different style transfer approaches integrated into our DASK framework. While both CoP and the VGG-based DSTN enhance model performance, our approach yields more significant improvements. This superiority is attributed to the limitations of CoP, which lacks context constraints, and DSTN, which tends to overfit the training data. In contrast, our AKPNet-based style transfer approach generates more realistic data, thereby promoting the model's lifelong learning capacity.

Additionally, Fig~\ref{fig:vgg-reconstruction}visualizes the images generated by CoP, DSTN, and our method. The visual comparison reveals that CoP produces overexposed images, while DSTN struggles to reconstruct the fine-grained details of the original images. By contrast, our method produces more realistic images, owing to the instance-adaptive style transfer prediction mechanism, which enhances the quality of the generated data and thus supports more effective lifelong learning.

\begin{figure}[t]
		\begin{center}
			\includegraphics[width=1.0\linewidth]{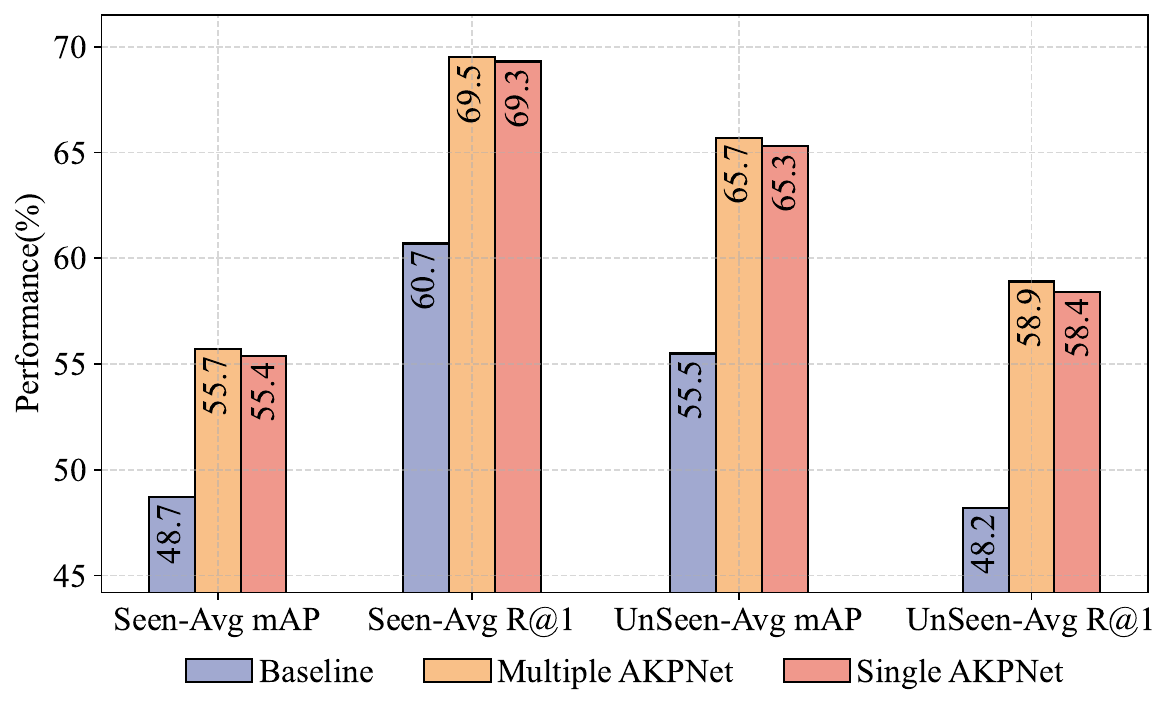}
			\caption{Ablation on different numbers of preserved historical AKPNet models. 
   }
			\label{fig:multi-akpnet}
		\end{center}    
	\end{figure}
\subsection{Preserving Multiple AKPNet Models}
In the proposed approach, only a single historical AKPNet model, $\boldsymbol{\mathrm{\Psi}}_{t-1}$, is retained to avoid the continuous increase in storage overhead. In this section, we investigate the scenario where all historical AKPNet models are preserved and utilized. Specifically, during training on $D_t$, the previous $t-1$ AKPNet models are randomly selected to generate old-style data. 
 The results presented in Fig~\ref{fig:multi-akpnet} indicate that utilizing a single $\boldsymbol{\mathrm{\Psi}}_{t-1}$ achieves comparable performance to preserving and using all historical AKPNet models. This outcome is attributed to the effectiveness of the distribution rehearsing mechanism, which strengthens the features learned from previous training steps. Consequently, the risk of forgetting caused by the acquisition of new features in subsequent training steps is significantly mitigated. Based on this observation, we adopt the strategy of preserving only a single  $\boldsymbol{\mathrm{\Psi}}_{t-1}$ by default.

\section{Details of the LReID Benchmark}
The LReID benchmark ~\cite{pu2021lifelong} consists of 12 existing ReID datasets. The detailed statistics of which are shown in Table~\ref{tab:datasets}. `Original Identities' denotes the identity numbers in the original datasets and `LReID Identities' denotes the selected identity numbers in the LReID benchmark. In the LReID benchmark, `LReID Identities' of all training datasets are set to 500 to mimic the identity-balanced conditions.
Note that CUHK-SYSU was initially proposed for the person search task and we preprocess it for the ReID task following~\cite{pu2021lifelong} where the ground-truth person bounding box annotation is used to crop individual-level images and a subset in which each identity contains at least 4 bounding boxes is selected.

\bibliography{aaai25}
\end{document}